\documentclass[10pt,letterpaper]{article}
\usepackage[top=0.85in,left=2.75in,footskip=0.75in]{geometry}

\usepackage{amsmath,amssymb}

\usepackage{changepage}

\usepackage{textcomp,marvosym}

\usepackage{cite}

\usepackage{nameref,hyperref}

\usepackage[right]{lineno}
\usepackage{graphicx}%
\usepackage{multirow}%
\usepackage{amsmath,amssymb,amsfonts}%
\usepackage{amsthm}%
\usepackage{mathrsfs}%
\usepackage[title]{appendix}%
\usepackage{xcolor}%
\usepackage{textcomp}%
\usepackage{manyfoot}%
\usepackage{booktabs}%
\usepackage{algorithm}%
\usepackage{algorithmicx}%
\usepackage{algpseudocode}%
\usepackage{listings}%
\usepackage{multirow}
\usepackage{booktabs}
\usepackage{xcolor}
\usepackage{tikz}
\usepackage{tabularx}
\usepackage{amsmath,amssymb}
\usepackage{subcaption}
\usepackage{pgfplots}

\pgfplotsset{compat=1.18}
\usepgfplotslibrary{groupplots}

\usetikzlibrary{positioning,calc,fit,arrows.meta,shapes.geometric,shapes.symbols,backgrounds}

\usepackage[nopatch=eqnum]{microtype}
\DisableLigatures[f]{encoding = *, family = * }

\usepackage[table]{xcolor}

\usepackage{array}

\newcolumntype{+}{!{\vrule width 2pt}}

\newlength\savedwidth

\definecolor{greenfill}{HTML}{DCEFD8}
\definecolor{redfill}{HTML}{E9DDF2}
\definecolor{sscanclasszero}{HTML}{0072B2}
\definecolor{sscanclassone}{HTML}{D55E00}
\definecolor{sscansupportfill}{HTML}{EAF3F8}
\definecolor{sscanqueryfill}{HTML}{FFF1E8}
\definecolor{sscannormfill}{HTML}{FFF4CC}
\definecolor{sscanencoderfill}{HTML}{E8EEF2}
\definecolor{sscanprototypefill}{HTML}{DCEFD8}
\definecolor{sscancanfill}{HTML}{E9DDF2}
\definecolor{sscanpanelgray}{HTML}{666666}

\usepackage[aboveskip=1pt,labelfont=bf,labelsep=period,justification=raggedright,singlelinecheck=off]{caption}

\makeatletter
\renewcommand{\@biblabel}[1]{\quad#1.}
\makeatother

\usepackage{lastpage,fancyhdr,graphicx}
\usepackage{epstopdf}
\fancyheadoffset[L]{2.25in}
\fancyfootoffset[L]{2.25in}
\begin{document}
\vspace*{0.2in}

\begin{flushleft}
{\Large
\textbf\newline{Few-Shot Learning for Personalised Automated Pain Assessment} 
}
\newline
\\
Heinke Hihn\textsuperscript{1,*},
Ibrahim Eisawy\textsuperscript{1},
Patrick Thiam\textsuperscript{2,3},
Hans A. Kestler\textsuperscript{3},
Friedhelm Schwenker\textsuperscript{2},
\\
\bigskip
\textbf{1} IU International University of Applied Sciences, Berlin, Germany
\\
\textbf{2} Institute of Neural Information Processing, Ulm University, Ulm, Germany
\\
\textbf{3} Institute of Medical Systems Biology, Ulm University, Ulm, Germany
\\
\bigskip

* heinke.hihn@iu.org

\end{flushleft}
\section*{Abstract}
Pain perception varies substantially across individuals, making it difficult for population-based classifiers to generalise across all subjects in a dataset. One way to account for subject variability is to train personalised classifiers. In this work, we evaluate Few-Shot Learning, a sub-area of Meta-Learning, as an approach to personalisation in automated pain assessment. We re-interpret the shift from population-level to subject-level evaluation as a task-domain shift, where the observed classes remain fixed but the target subject changes. We evaluate our method on the BioVid Pain Database, the SenseEmotion Database, and the PainMonit Experimental Dataset (PMED), reaching 85.75\% and 35.49\% accuracy on BioVid and 82.37\% and 41.88\% on SenseEmotion in the binary and multi-class settings under a Leave-One-Subject-Out CV protocol respectively, and 90.47\% on PMED, for which only a binary benchmark exists. Using samples to implement $k$-shot conditioning, the accuracies can be improved to 86.25\%, 40.06\%, 83.43\%, 44.08\%, and 91.25\%, respectively. To further evaluate the effects and robustness of our method, we provide additional ablation experiments and investigate the personalisation effects. Our results suggest that support-conditioned few-shot adaptation can improve average performance under inter-subject variability.

\section*{Author summary}
This study investigates how Meta-Learning can be used to personalise population-level classifiers for Automated Pain Assessment to reduce inter-subject variance. This is still a major issue in subject-based data analysis. To this end, this study proposes a Few-Shot Learning mechanism that calibrates a model using a few labelled samples. The method is evaluated using three multimodal datasets and investigates personalisation benefits on a class level and on the subject level. The analysis revealed that not all classes and not all subjects benefit equally from personalisation: the intermediate classes, i.e., not the extreme no pain and highest pain, benefit the most from personalisation. The results of the study suggest that the Few-Shot Learning based approach offers an efficient way to build personalised classifiers.


\section{Introduction}
\label{sec:intro}
Pain is a highly personal, private, and subjective emotion~\cite{werner2019automatic} that is characterised by severity, location, duration, and intensity. In addition, the pain experience is modulated by several personal factors, including cognition, past experiences, and the current situation~\cite{werner2019automatic}. As a result, the same pain stimulus can cause different reactions, or no reaction at all. These factors make it difficult to reliably assess pain leading to self-reporting tools such as the Visual Analogue Scale (VAS)~\cite{hawker2011measures} being the most common method. In cases where the patient is not able to articulate their pain, caregivers rely on visible reactions of the body and bio-physical signals, if available. In such cases, automated systems could provide decision support to assist physicians in determining the pain level a patient is in, which can help improve patient well-being and recovery.

To advance towards such systems, researchers have collected data in controlled experiments, where healthy subjects were exposed to varying levels of pain stimuli and their bodily responses were recorded \cite{Walter2013BioVid,velana2016senseemotion,gouverneur2024experimental}. Such recordings range from bio-physical signals such as Electrodermal Activity (EDA), Electrocardiogram (ECG), and Electromyography (EMG) to video recordings of their facial expressions. These datasets have led to a set of automated pain assessment methods, ranging from hand-made features~\cite{kachele2016methods} to deep learning~\cite{Thiam2021DeepPhys,thiam2019exploring,gkikas2025efficient}. 

The highly subjective nature of pain leads to high inter-subject variability of classifiers trained on such generic population data,  which is a widely documented challenge in automated pain systems~\cite{khan2025systematic,werner2019automatic}. To overcome this challenge, automated pain assessment requires models that generalise across individuals, because pain expressions and physiological responses vary strongly between subjects. We group the approaches to this problem into
four overlapping conceptual categories:

\textbf{(i) Preprocessing-level personalisation:} Inter-subject variability is handled before or during modelling by
making physiological responses more comparable across subjects. Strategies include subject-specific feature standardisation,
individual pain-threshold calibration, and subject-aware normalisation~\cite{ji2024unraveling,pouromran2021exploration,werner2016automatic}.

\textbf{(ii) Model-level personalisation}: The model is explicitly adapted to the individual rather than assuming one
universal pain-response pattern. Ideas include subject-dependent models, personalised feature fusion, dynamic attention,
and personal pain sensitivity representations~\cite{jiang2024personalized,pouromran2021exploration}.

\textbf{(iii) Population-plus-individual modelling:} Individual differences are not treated as noise but as meaningful
structure that can be statistically or computationally modelled. Approaches include multi-task learning, person-specific
outputs, subject profiles, clustering, and mixed-effects models~\cite{bourou2018video,lopez2017multi,kachele2016methods,gozzi2024unraveling}.

\textbf{(iv) Generalisation-focused evaluation and adaptation}: The central question is whether a model trained on
source subjects generalises to a held-out subject. Strategies span transfer learning, uncertainty-guided personalisation,
and Few-Shot-style adaptation~\cite{werner2016automatic,gozzi2024unraveling,jiang2024personalized,ji2024unraveling}.

Several machine learning paradigms exist that aim to build adaptive classifiers, such as Meta-Learning~\cite{hospedales2021meta,hihn2020hierarchical,hihn2020specialization}, Continual Learning~\cite{wang2024comprehensive,hihn2021mixture,hihn2023hierarchically,hihn2024online}, and Self-Supervised Learning~\cite{gui2024survey}. While Self-Supervised Learning has seen recent applications to pain assessment~\cite {thiam2021multi} and processing of bio-physical signals~\cite{del2023applications}, as well as Continual Learning in affective computing~\cite{thuseethan2021deep}, only a few studies have evaluated Meta-Learning for automated pain assessment.

To close this gap, we evaluate Meta-Learning as a candidate approach for devising personalised automated pain assessment systems to reduce inter-subject variability. In contrast to standard supervised learning, Meta-Learning aims to adapt the learning algorithm itself, so that instead of learning from scratch, it leverages previously acquired knowledge to adapt quickly to new problems. Specifically, we focus on Few-Shot Learning, which is designed for settings in which a model must adapt to a new task or individual from only a small number of labelled examples~\cite{hihn2020specialization,hihn2020hierarchical,wang2020generalizing}. Considering that collecting data in pain assessment experiments is expensive and we typically have 8~\cite{gouverneur2024experimental} to 20~\cite{Walter2013BioVid} samples per class and per subject, Few-Shot Learning matches the combination of subject variability and limited labelled data that characterises personalised automated pain assessment.

To systematically investigate Few-Shot Learning in the context of personalised automated pain assessment, we aim to answer the following
research questions:

\begin{enumerate}
  \item \textit{To what extent can personalisation improve automated pain assessment under inter-individual variability?}
  \item \textit{Do the benefits of personalisation vary across individuals?}
  \item \textit{Do the benefits of personalisation vary across pain levels?}
\end{enumerate}

To answer these research questions, this work makes the following contributions to automated pain assessment:

\begin{enumerate}
    \item we propose an approach utilising Few-Shot Learning to build personalised classifiers,
    \item we introduce a Cross Alignment Network allowing us to efficiently score samples against class-prototypes, and
    \item we introduce a Zero-Shot extension of our method, allowing us to evaluate samples without requiring labelled support samples.
\end{enumerate}

This article is structured as follows: first, we discuss related work in Section~\ref{sec:relatedwork}. We introduce our
core method in Section~\ref{sec:method}, evaluate it in Section~\ref{sec:results}, and conclude with a discussion in
Section~\ref{sec:discussion}.

\section{Related Work}
The method our work introduces is related to three sub-fields: methods that incorporate Few-Shot Learning into Leave-One-Subject-Out (LOSO) evaluation protocols, prototype-based methods for the automated assessment of pain, and personalisation in automated pain assessment. In the following, we discuss studies that fall into these categories.
\label{sec:relatedwork}
\subsection{Combining Few-Shot Learning with LOSO Evaluation}
Few-shot learning has been combined with Leave-One-Subject-Out evaluation protocols in several cross-subject
EEG~\cite{an2023dual}, physiological signal~\cite{wang2026msfsnet}, and human activity recognition studies~\cite{belal2025fsid}.
In such settings, the model is typically trained or meta-trained on all source subjects, while one subject is held out
as the target subject. During evaluation, a small labelled support set from the held-out subject is used for adaptation
or calibration when the evaluation protocol includes target-subject support samples, and the remaining samples from the
same subject form the query/test set. This differs from strict LOSO evaluation, where no target-subject samples are
available during training or adaptation.

Few-shot learning has also been explored in automated pain assessment, with fewer published studies than in related
areas such as EEG emotion recognition or human activity recognition. Existing work includes meta-learning approaches for facial-expression-based pain detection, such as Model Agnostic Meta-Learning-based adaptation on the UNBC-McMaster pain archive~\cite{rathee2022pain}, as well as Few-Shot or prompt-based approaches for neonatal and infant pain assessment from visual or acoustic data~\cite{buzuti2024generative,mccofie2025few,el2026pandia}. These studies apply Few-Shot learning to pain-assessment settings with limited labelled data and inter-individual variability.

Closest to our work is the approach of Rathee et al.~\cite{rathee2022pain}, where the authors utilise a modified Model-Agnostic Meta-Learning approach, which can learn to detect pain intensity from facial expressions. They evaluated the proposed model on the UNBC-McMaster pain archive database and report higher accuracy for 1-shot and 5-shot learning than their baselines.

In these studies, Few-Shot learning is used to address recurring challenges in pain assessment, especially limited
labelled data, inter-individual variability, and the need for rapid adaptation to new subjects or patients. These studies motivate evaluating Few-Shot learning for automated pain assessment.

\subsection{Prototype-Based Classification in Pain Assessment}
Prototype-based methods represent each class or cluster using one or more representative reference points, known as prototypes, and classify new samples based on their similarity or distance to these prototypes. This paradigm has been applied to pain assessment across multiple data modalities, including fMRI-based pain decoding \cite{anter2023rlwoa}, clustering of chronic pain patient records \cite{khan2026chronic}, and knowledge transfer between phasic and tonic pain stimuli \cite{ricken2023pain}. These methods provide interpretable and flexible prototypes that are learned directly from the data. The prototypes naturally capture the heterogeneity of patient populations, and can serve as representative anchors for transferring knowledge across related pain conditions.

Anter et al.~\cite{anter2023rlwoa} propose a model for decoding pain intensity from fMRI data. Their method uses both a reinforcement learning-enhanced whale optimisation algorithm (RLWOA) to identify discriminative pain-related fMRI features and a self-organizing fuzzy logic (SOFL) classifier to predict different pain intensities. The SOFL classifier automatically identifies data prototypes and constructs interpretable fuzzy rules without requiring predefined membership functions or assumptions about the underlying data distribution. The RLWOA-SOFL model also supports multi-class pain intensity classification, which is particularly relevant for clinical pain assessment, where pain is typically evaluated across up to 10 intensity levels.

Khan et al.~\cite{khan2026chronic} apply $k$-prototypes clustering as an unsupervised learning technique to analyse chronic pain patient data and identify clinically meaningful clusters. They adopt $k$-prototypes because it is a partitioning-based method well suited to mixed-type or heterogeneous data. Unlike $k$-means or $k$-modes, which are restricted to numerical or categorical variables respectively, $k$-prototypes can handle both types simultaneously. The algorithm defines a set of virtual points, or prototypes, as the centre of the clusters. For numerical variables, these prototypes are represented by their mean values, and for categorical variables, these prototypes are represented by mode values. The algorithm then measures dissimilarity between two mixed-type objects A and B by combining a squared Euclidean distance over numerical features with a simple matching count over categorical ones. In their study, these prototype-based clusters are used to examine whether patient-reported chronic pain data align with established nociceptive, neuropathic, and nociplastic pain mechanisms.

Ricken et al.~\cite{ricken2023pain} apply prototype-based knowledge transfer from short-term (phasic) to long-term (tonic) pain stimuli using physiological signals of the X-ITE Pain Database. A classification model is trained on the phasic domain, and class-specific prototypes (one for the no-pain and one for the pain class) are computed as feature-average samples of that domain. Each tonic sample is then partitioned into equally sized segments matching the length of the phasic samples, and the distance between every segment and each phasic prototype is measured. The segment with the smallest distance to the prototypes is selected and passed to the classifier to label the entire tonic sample. They evaluate a range of distance metrics for this segment selection, including Euclidean, City-Block, Chebyshev, Canberra, Bray–Curtis, and Wasserstein. The authors also compare this distance-based segment selection to a naive approach that feeds unmodified tonic stimuli to the phasic-trained model without adaptation. The distance-based selection of a single segment improves classification performance across most metrics; for example, on the trapezius signal (TRA) in the electric pain domain, it raised average accuracy from 66.0$\%$ to 68.0$\%$, and on the electrodermal activity signal in the thermal domain, from 53.2$\%$ to 59.6$\%$.

\subsection{Personalisation in Automated Pain Assessment}
Sharafi et al.~\cite{sharafi2025disentangled} introduce the Disentangled Source-Free Domain Adaptation (DSFDA) method,
which addresses inter-subject variability and the challenge of adapting models with missing target
expression data by leveraging neutral control videos for end-to-end generation and adaptation. Their setting is subject-based: a source model is pre-trained on all classes from the source subjects, then adapted to each target subject individually using only a short neutral control video, while testing covers both neutral and pain expressions. To compensate for the missing classes, DSFDA learns to disentangle expression-related from identity-related features and generates the absent non-neutral images for the target subject, combining data generation and adaptation in a single end-to-end stage. A self-supervision strategy further reconstructs target images preserving the same identity and source expression. The approach is evaluated on the BioVid and UNBC-McMaster pain assessment datasets, outperforming state-of-the-art source-free adaptation methods, which degrade sharply when only neutral target data is available for adaptation.

Jiang et al.~\cite{jiang2024personalized} propose a personalised neural network for time-continuous pain detection. The model addresses the inter-subject variability in the data by combining personalised feature fusion based on pain sensitivity with a dynamic feature attention mechanism. Pain sensitivity accounts for individual differences in pain perception, while the dynamic feature attention mechanism enables the model to focus on the most informative physiological features over time and better capture varying pain response patterns. The approach is evaluated on both the authors' SpaExp and the BioVid Heat Pain dataset. It demonstrates improved cross-dataset generalisation across different pain protocols and shows competitive performance in pain intensity classification.

Fang et al.~\cite{fang2022atlas} proposed the Adaptive Transfer Learning-based Pain Assessment System (ATLAS) to improve automatic pain assessment for new patients whose data are unlabelled and may differ from the data used to train the model. ATLAS applies Transfer Component Analysis (TCA) to reduce the distribution gap between the labelled source data and the new patient’s physiological signals, and adaptively combines the resulting personalised model with a general pretrained model. The framework was evaluated on the BioVid database using ECG, skin conductance, and electromyography signals. The results showed the approach outperformed three existing traditional machine learning-based approaches and achieved an average accuracy only 2$\%$ below that of supervised methods using labelled test data.

Thiam et al.~\cite{thiam2017hierarchical} propose a personalised, participant-independent pain recognition system based on the video channel to improve automatic pain assessment for an unseen participant whose facial responses may differ from those of the participants used to train the model. Several complementary facial descriptors, including landmark distances, head pose, LBP-TOP, LGBP-TOP, and PHOG, are each classified by a separate Random Forest and combined through a two-stage hierarchical fusion scheme. Rather than training on the entire annotated dataset, the system uses the Hausdorff distance and two variants based on the average and median of the distances between the points of two sets to measure the similarity between the unseen participant's data and that of each training participant, and trains the model only on the most similar subset. The framework is evaluated on data collected within the SenseEmotion project of 40 participants undergoing thermal pain stimulation, for the task of distinguishing no stimulation from maximum pain. The results show that selecting around 30 participants outperformed the baseline that used all available data. However, the gains were small on the two datasets corresponding to the left and right forearm stimulation sites (65.7$\%$ to 66.8$\%$ and 67.6$\%$ to 67.8$\%$, respectively), with the modest overall accuracy attributed to participants who displayed little visible facial reaction to the stimuli.

\section{Method}
\label{sec:method}
In the following, we will introduce the datasets we use for evaluation (Section~\ref{sec:datasets}) and the building blocks that comprise our method: $(i)$ how we re-frame automated pain assessment as subject-as-a-task prototype-based Few-Shot learning (Section~\ref{sec:few_shot_construction}), $(ii)$ how we apply cross-modal feature fusion to incorporate EDA and ECG signals (Section~\ref{subsec:crossmod_fusion}, $(iii)$ our proposed cross-attention scoring network used to compare support prototypes with query inputs (Section~\ref{subsec:can_scoring}), and $(iv)$ how we propose to
turn our method into a Zero-Shot classifier that does not require any labelled samples from the held-out subject
(Section~\ref{sec:prototype_bank}). 

\subsection{Datasets}
\label{sec:datasets}
\begin{figure}[t]
    \centering
    \includegraphics[]{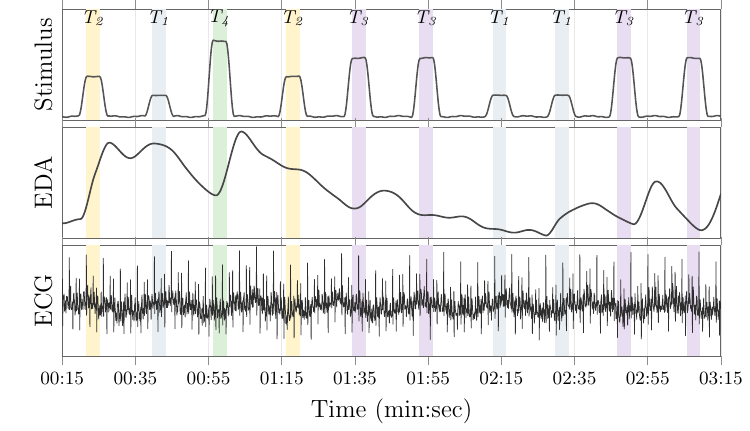}
    \caption{The data in the experiments has been generated by subjecting subjects to various levels of pain (annotated as $T_i$) and a baseline of no pain and recording their bodily response by various bio-physical signals such as EDA, EMG, and ECG. This creates time series of signals annotated by the experienced pain level, thus enabling us to train models to predict the pain based on the signals.}
    \label{fig:data_creation}
\end{figure}

\subsubsection{The BioVid Pain Dataset}
\label{sec:biovid}

We evaluate our method using the BioVid Heat Pain Database (Part A)~\cite{Walter2013BioVid}. It consists of 87
participants who were exposed to four levels of increasing thermal pain, indicated as $T_1$, $T_2$, $T_3$, and $T_4$.
During the experiment, video streams, EDA, ECG, and EMG signals were recorded for each participant, enabling us to map
their bodily reactions to the pain stimulus. Each pain level was elicited 20 times in random order, where each elicitation
lasted $4s$ and was followed by a recovery phase ($T_0$ with 32°C) between $8$ and $12s$. Thus, in total the dataset
consists of $20 \times 5 \times 87 = 8,700$ samples. We use the data augmentation described by Thiam et al.~\cite{Thiam2021DeepPhys},
which gives us $87,000$ samples in total. For our experiments, we rely only on the EDA and the ECG signals, following
prior physiological pain-assessment work that emphasises EDA~\cite{Lu2023PainAttnNet,Aziz2024MDNet,li2024automatic,shi2022tree,CrossModTransformer2025,aslam2025learning,thiam2025dealing}.

\subsubsection{The SenseEmotion Dataset}
\label{sec:senseemotion}

Additionally, we assess our method on the SenseEmotion dataset~\cite{velana2016senseemotion}. The dataset consists of 45 healthy participants, each exposed to three increasing heat pain levels, indicated as T1, T2, T3. For this experiment, the signals (biopotentials: SCL/EDA, ECG, trapezius EMG, respiration), plus video (facial expressions, skeleton, head pose), were recorded. These signals enabled us to map body reactions to a certain pain level. Each pain level was applied 30 times for $4s$ in random order, with randomised pauses ($T_0$ with 32°C) of $8-13s$ between stimuli. This dataset consists of $3 \times 30 \times 45 = 4050$ pain-stimulus samples. For our experiments, we only use EDA and ECG signals for pain assessment, following prior physiological pain-assessment work ~\cite{Lu2023PainAttnNet,Aziz2024MDNet,li2024automatic,shi2022tree,CrossModTransformer2025,aslam2025learning,thiam2025dealing}.

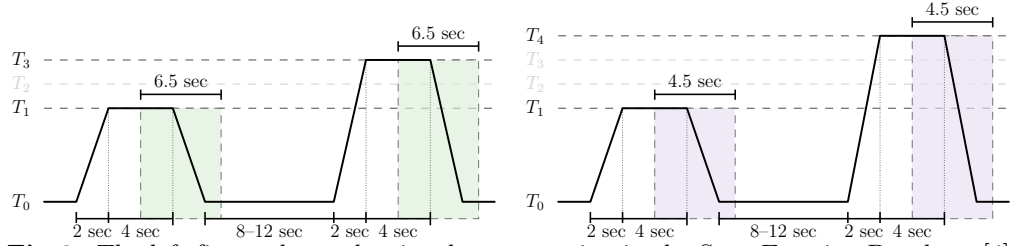
\begin{figure}[t]
    \centering
    \begin{minipage}[t]{0.49\textwidth}
        \centering
        \resizebox{\linewidth}{!}{%
            \begin{tikzpicture}[
    x=0.9cm,
    y=1.35cm,
    signal/.style={
        black,
        line width=1.25pt,
        line cap=round,
        line join=round
    },
    guide/.style={
        black!70,
        line width=0.6pt,
        dash pattern=on 5pt off 5pt
    },
    guide faded/.style={
        black!18,
        line width=0.6pt,
        dash pattern=on 5pt off 5pt
    },
    window border/.style={
        black!55,
        line width=0.6pt,
        dash pattern=on 5pt off 4pt
    },
    event line/.style={
        black!55,
        line width=0.6pt,
        densely dotted
    },
    duration/.style={
        black,
        line width=1pt
    },
    level label/.style={
        anchor=east,
        font=\large
    },
    time label/.style={
        font=\large,
        inner sep=1pt,
        fill=white
    }
]

\def\TZero{0}
\def\TOne{1.55}
\def\TTwo{1.95}
\def\TThree{2.35}

\def\Bottom{-0.28}

\def\SignalStart{0.55}

\def\RiseOneStart{1.35}
\def\PlateauOneStart{2.15}
\def\FallOneStart{3.75}
\def\FallOneEnd{4.55}

\def\RiseTwoStart{7.75}
\def\PlateauTwoStart{8.55}
\def\FallTwoStart{10.15}
\def\FallTwoEnd{10.95}

\def\SignalEnd{11.75}

\def\WindowOneStart{2.95}
\def\WindowOneEnd{4.95}

\def\WindowTwoStart{9.35}
\def\WindowTwoEnd{11.35}

\draw[guide]
    (0.55,\TOne) -- (11.75,\TOne);

\draw[guide faded]
    (0.55,\TTwo) -- (11.75,\TTwo);

\draw[guide]
    (0.55,\TThree) -- (11.75,\TThree);

\path[
    fill=greenfill,
    fill opacity=0.65
]
    (\WindowOneStart,\Bottom)
    rectangle
    (\WindowOneEnd,\TOne);

\draw[window border]
    (\WindowOneStart,\Bottom)
    rectangle
    (\WindowOneEnd,\TOne);

\path[
    fill=greenfill,
    fill opacity=0.65
]
    (\WindowTwoStart,\Bottom)
    rectangle
    (\WindowTwoEnd,\TThree);

\draw[window border]
    (\WindowTwoStart,\Bottom)
    rectangle
    (\WindowTwoEnd,\TThree);

\draw[event line]
    (\PlateauOneStart,\Bottom)
    -- (\PlateauOneStart,\TOne);

\draw[event line]
    (\FallOneStart,\Bottom)
    -- (\FallOneStart,\TOne);

\draw[event line]
    (\PlateauTwoStart,\Bottom)
    -- (\PlateauTwoStart,\TThree);

\draw[event line]
    (\FallTwoStart,\Bottom)
    -- (\FallTwoStart,\TThree);

\draw[signal]
    (\SignalStart,\TZero)
    -- (\RiseOneStart,\TZero)
    -- (\PlateauOneStart,\TOne)
    -- (\FallOneStart,\TOne)
    -- (\FallOneEnd,\TZero)
    -- (\RiseTwoStart,\TZero)
    -- (\PlateauTwoStart,\TThree)
    -- (\FallTwoStart,\TThree)
    -- (\FallTwoEnd,\TZero)
    -- (\SignalEnd,\TZero);

\node[level label]
    at (0.35,\TZero) {$T_0$};

\node[level label]
    at (0.35,\TOne) {$T_1$};

\node[level label, text=black!18]
    at (0.35,\TTwo) {$T_2$};

\node[level label]
    at (0.35,\TThree) {$T_3$};


\draw[duration]
    (\WindowOneStart,{ \TOne + 0.14 })
    -- (\WindowOneStart,{ \TOne + 0.31 });

\draw[duration]
    (\WindowOneEnd,{ \TOne + 0.14 })
    -- (\WindowOneEnd,{ \TOne + 0.31 });

\draw[duration]
    (\WindowOneStart,{ \TOne + 0.23 })
    -- (\WindowOneEnd,{ \TOne + 0.23 });

\node[
    font=\large,
    anchor=south
]
    at ({(\WindowOneStart+\WindowOneEnd)/2},{ \TOne + 0.26 })
    {6.5 sec};

\draw[duration]
    (\WindowTwoStart,{ \TThree + 0.14 })
    -- (\WindowTwoStart,{ \TThree + 0.31 });

\draw[duration]
    (\WindowTwoEnd,{ \TThree + 0.14 })
    -- (\WindowTwoEnd,{ \TThree + 0.31 });

\draw[duration]
    (\WindowTwoStart,{ \TThree + 0.23 })
    -- (\WindowTwoEnd,{ \TThree + 0.23 });

\node[
    font=\large,
    anchor=south
]
    at ({(\WindowTwoStart+\WindowTwoEnd)/2},{ \TThree + 0.26 })
    {6.5 sec};


\draw[duration]
    (\RiseOneStart,\Bottom)
    -- (\PlateauOneStart,\Bottom);

\draw[duration]
    (\RiseOneStart,{-0.39})
    -- (\RiseOneStart,{-0.17});

\draw[duration]
    (\PlateauOneStart,{-0.39})
    -- (\PlateauOneStart,{-0.17});

\node[time label, anchor=north]
    at (1.75,-0.35)
    {2 sec};

\draw[duration]
    (\PlateauOneStart,\Bottom)
    -- (\FallOneStart,\Bottom);

\draw[duration]
    (\FallOneStart,{-0.39})
    -- (\FallOneStart,{-0.17});

\node[time label, anchor=north]
    at (2.95,-0.35)
    {4 sec};

\draw[duration]
    (\FallOneEnd,\Bottom)
    -- (\RiseTwoStart,\Bottom);

\draw[duration]
    (\FallOneEnd,{-0.39})
    -- (\FallOneEnd,{-0.17});

\draw[duration]
    (\RiseTwoStart,{-0.39})
    -- (\RiseTwoStart,{-0.17});

\node[time label, anchor=north]
    at (6.15,-0.35)
    {8--12 sec};

\draw[duration]
    (\RiseTwoStart,\Bottom)
    -- (\PlateauTwoStart,\Bottom);

\draw[duration]
    (\PlateauTwoStart,{-0.39})
    -- (\PlateauTwoStart,{-0.17});

\node[time label, anchor=north]
    at (8.15,-0.35)
    {2 sec};

\draw[duration]
    (\PlateauTwoStart,\Bottom)
    -- (\FallTwoStart,\Bottom);

\draw[duration]
    (\FallTwoStart,{-0.39})
    -- (\FallTwoStart,{-0.17});

\node[time label, anchor=north]
    at (9.35,-0.35)
    {4 sec};

\end{tikzpicture}
        }
    \end{minipage}
    \hfill
    \begin{minipage}[t]{0.49\textwidth}
        \centering
        \resizebox{\linewidth}{!}{%
\begin{tikzpicture}[
    x=0.9cm,
    y=1.35cm,
    signal/.style={
        black,
        line width=1.25pt,
        line cap=round,
        line join=round
    },
    guide/.style={
        black!70,
        line width=0.6pt,
        dash pattern=on 5pt off 5pt
    },
    guide faded/.style={
        black!18,
        line width=0.6pt,
        dash pattern=on 5pt off 5pt
    },
    window border/.style={
        black!55,
        line width=0.6pt,
        dash pattern=on 5pt off 4pt
    },
    event line/.style={
        black!55,
        line width=0.6pt,
        densely dotted
    },
    duration/.style={
        black,
        line width=1pt
    },
    level label/.style={
        anchor=east,
        font=\large
    },
    time label/.style={
        font=\large,
        inner sep=1pt,
        fill=white
    }
]

\def\TZero{0}
\def\TOne{1.55}
\def\TTwo{1.95}
\def\TThree{2.35}
\def\TFour{2.75}

\def\Bottom{-0.28}

\def\SignalStart{0.55}

\def\RiseOneStart{1.35}
\def\PlateauOneStart{2.15}
\def\FallOneStart{3.75}
\def\FallOneEnd{4.55}

\def\RiseTwoStart{7.75}
\def\PlateauTwoStart{8.55}
\def\FallTwoStart{10.15}
\def\FallTwoEnd{10.95}

\def\SignalEnd{11.75}

\def\WindowOneStart{2.95}
\def\WindowOneEnd{4.95}

\def\WindowTwoStart{9.35}
\def\WindowTwoEnd{11.35}

\draw[guide]
    (0.55,\TOne) -- (11.75,\TOne);

\draw[guide faded]
    (0.55,\TTwo) -- (11.75,\TTwo);

\draw[guide faded]
    (0.55,\TThree) -- (11.75,\TThree);

\draw[guide]
    (0.55,\TFour) -- (11.75,\TFour);

\path[
    fill=redfill,
    fill opacity=0.65
]
    (\WindowOneStart,\Bottom)
    rectangle
    (\WindowOneEnd,\TOne);

\draw[window border]
    (\WindowOneStart,\Bottom)
    rectangle
    (\WindowOneEnd,\TOne);

\path[
    fill=redfill,
    fill opacity=0.65
]
    (\WindowTwoStart,\Bottom)
    rectangle
    (\WindowTwoEnd,\TFour);

\draw[window border]
    (\WindowTwoStart,\Bottom)
    rectangle
    (\WindowTwoEnd,\TFour);

\draw[event line]
    (\PlateauOneStart,\Bottom)
    -- (\PlateauOneStart,\TOne);

\draw[event line]
    (\FallOneStart,\Bottom)
    -- (\FallOneStart,\TOne);

\draw[event line]
    (\PlateauTwoStart,\Bottom)
    -- (\PlateauTwoStart,\TFour);

\draw[event line]
    (\FallTwoStart,\Bottom)
    -- (\FallTwoStart,\TFour);

\draw[signal]
    (\SignalStart,\TZero)
    -- (\RiseOneStart,\TZero)
    -- (\PlateauOneStart,\TOne)
    -- (\FallOneStart,\TOne)
    -- (\FallOneEnd,\TZero)
    -- (\RiseTwoStart,\TZero)
    -- (\PlateauTwoStart,\TFour)
    -- (\FallTwoStart,\TFour)
    -- (\FallTwoEnd,\TZero)
    -- (\SignalEnd,\TZero);

\node[level label]
    at (0.35,\TZero) {$T_0$};

\node[level label]
    at (0.35,\TOne) {$T_1$};

\node[level label, text=black!18]
    at (0.35,\TTwo) {$T_2$};

\node[level label, text=black!18]
    at (0.35,\TThree) {$T_3$};

\node[level label]
    at (0.35,\TFour) {$T_4$};


\draw[duration]
    (\WindowOneStart,{ \TOne + 0.14 })
    -- (\WindowOneStart,{ \TOne + 0.31 });

\draw[duration]
    (\WindowOneEnd,{ \TOne + 0.14 })
    -- (\WindowOneEnd,{ \TOne + 0.31 });

\draw[duration]
    (\WindowOneStart,{ \TOne + 0.23 })
    -- (\WindowOneEnd,{ \TOne + 0.23 });

\node[
    font=\large,
    anchor=south
]
    at (3.95,{ \TOne + 0.26 })
    {4.5 sec};

\draw[duration]
    (\WindowTwoStart,{ \TFour + 0.14 })
    -- (\WindowTwoStart,{ \TFour + 0.31 });

\draw[duration]
    (\WindowTwoEnd,{ \TFour + 0.14 })
    -- (\WindowTwoEnd,{ \TFour + 0.31 });

\draw[duration]
    (\WindowTwoStart,{ \TFour + 0.23 })
    -- (\WindowTwoEnd,{ \TFour + 0.23 });

\node[
    font=\large,
    anchor=south
]
    at (10.35,{ \TFour + 0.26 })
    {4.5 sec};


\draw[duration]
    (\RiseOneStart,\Bottom)
    -- (\PlateauOneStart,\Bottom);

\draw[duration]
    (\RiseOneStart,{-0.39})
    -- (\RiseOneStart,{-0.17});

\draw[duration]
    (\PlateauOneStart,{-0.39})
    -- (\PlateauOneStart,{-0.17});

\node[time label, anchor=north]
    at (1.75,-0.35)
    {2 sec};

\draw[duration]
    (\PlateauOneStart,\Bottom)
    -- (\FallOneStart,\Bottom);

\draw[duration]
    (\FallOneStart,{-0.39})
    -- (\FallOneStart,{-0.17});

\node[time label, anchor=north]
    at (2.95,-0.35)
    {4 sec};

\draw[duration]
    (\FallOneEnd,\Bottom)
    -- (\RiseTwoStart,\Bottom);

\draw[duration]
    (\FallOneEnd,{-0.39})
    -- (\FallOneEnd,{-0.17});

\draw[duration]
    (\RiseTwoStart,{-0.39})
    -- (\RiseTwoStart,{-0.17});

\node[time label, anchor=north]
    at (6.15,-0.35)
    {8--12 sec};

\draw[duration]
    (\RiseTwoStart,\Bottom)
    -- (\PlateauTwoStart,\Bottom);

\draw[duration]
    (\PlateauTwoStart,{-0.39})
    -- (\PlateauTwoStart,{-0.17});

\node[time label, anchor=north]
    at (8.15,-0.35)
    {2 sec};

\draw[duration]
    (\PlateauTwoStart,\Bottom)
    -- (\FallTwoStart,\Bottom);

\draw[duration]
    (\FallTwoStart,{-0.39})
    -- (\FallTwoStart,{-0.17});

\node[time label, anchor=north]
    at (9.35,-0.35)
    {4 sec};

\end{tikzpicture}

        }
    \end{minipage}

    \caption{The left figure shows the signal segmentation in the SenseEmotion Database \cite{velana2016senseemotion}, where experiments have been carried out on $6.5s$ windows with a $4s$ shift from the onset of the elicitation in four pain levels $T_0, ..., T_3$. On the right, we show how in the BioVid Database \cite{Walter2013BioVid} the windows are of $4.5s$ with an $4s$ shift on five pain levels $T_0, ..., T_4$. Thus, on sampling rate of 256 Hz we have an input dimensionality $1,664$ of and $1,152$, respectively.}
    \label{fig:data_augmentation}
\end{figure}

\subsubsection{The PainMonit Dataset}
\label{sec:painmonit}
Next, we assess our method on the PainMonit Database (PMED). It consists of 52 healthy participants who were exposed to 5 levels of increasing thermal pain, denoted as NP, $P_1$, $P_2$, $P_3$, and $P_4$.
During the experiment, EDA, ECG, EMG, BVP, Resp, Temp, and Grip signals were recorded for each participant. These signals allow us to map the patient's body reactions to the pain stimulus. Each pain level was elicited 8 times for $10s$ in random order, with a recovery phase ($T_0$ at 32°C) of between $20$ and $30s$. Overall, the dataset consists of $6 \times \sim 8 \times 52 = 2495$. In our experiments, we consider only the EDA and the ECG signals for pain assessment, following the prior physiological pain-assessment work ~\cite{Lu2023PainAttnNet,Aziz2024MDNet,li2024automatic,shi2022tree,CrossModTransformer2025,aslam2025learning,thiam2025dealing}.

\subsection{Few-Shot Task Construction}
\label{sec:few_shot_construction}
\begin{figure}[t]
    \centering

\begingroup

\tikzset{
  sscan panel/.style={
    draw=sscanpanelgray,
    rounded corners=2pt,
    line width=0.7pt,
    inner sep=5pt
  },
  sscan set/.style={
    draw=black!55,
    rounded corners=2pt,
    minimum width=30mm,
    minimum height=21mm,
    inner sep=3pt,
    align=center
  },
  sscan support/.style={sscan set, fill=sscansupportfill},
  sscan query/.style={sscan set, fill=sscanqueryfill, dashed},
  sscan process/.style={
    draw=black!65,
    rounded corners=2pt,
    minimum width=25mm,
    minimum height=14mm,
    inner sep=3pt,
    align=center
  },
  sscan norm/.style={sscan process, fill=sscannormfill},
  sscan encoder/.style={sscan process, fill=sscanencoderfill},
  sscan prototype/.style={sscan process, fill=sscanprototypefill},
  sscan can/.style={sscan process, fill=sscancanfill},
  sscan output/.style={
    sscan process,
    fill=black!4,
    minimum width=23mm
  },
  sscan map/.style={
    draw=black!55,
    rounded corners=1pt,
    minimum width=20mm,
    minimum height=8mm,
    fill=white,
    align=center,
  },
  sscan flow/.style={
    -{Latex[length=2.2mm]},
    line width=0.8pt,
    draw=black!75
  },
  sscan stats/.style={
    -{Latex[length=2mm]},
    line width=0.75pt,
    draw=black!70,
    dashed
  },
  sscan note/.style={
    align=center,
    text=black!70
  },
  sscan sample zero/.style={
    circle,
    draw=sscanclasszero,
    fill=sscanclasszero!20,
    line width=0.9pt,
    minimum size=4.6mm,
    inner sep=0pt
  },
  sscan sample one/.style={
    rectangle,
    draw=sscanclassone,
    fill=sscanclassone!20,
    line width=0.9pt,
    minimum size=4.2mm,
    inner sep=0pt
  },
  sscan query zero/.style={sscan sample zero, dashed},
  sscan query one/.style={sscan sample one, dashed}
}

\newcommand{\sscansupportsamples}[1]{%
  \node[sscan sample zero] at ($(#1.center)+(-7mm,-2mm)$) {};
  \node[sscan sample zero] at ($(#1.center)+(-2.3mm,-2mm)$) {};
  \node[sscan sample one] at ($(#1.center)+( 2.3mm,-2mm)$) {};
  \node[sscan sample one] at ($(#1.center)+( 7mm,-2mm)$) {};
}

\newcommand{\sscanquerysamples}[1]{%
  \node[sscan query zero] at ($(#1.center)+(-7mm,-2mm)$) {?};
  \node[sscan query zero] at ($(#1.center)+(-2.3mm,-2mm)$) {?};
  \node[sscan query one] at ($(#1.center)+( 2.3mm,-2mm)$) {?};
  \node[sscan query one] at ($(#1.center)+( 7mm,-2mm)$) {?};
}

\begin{tikzpicture}[font=\normalsize, node distance=5mm and 7mm]

\node[sscan support] (support) {};
\node[font=\footnotesize\bfseries, anchor=north] at
  ($(support.north)+(0,-1.5mm)$) {Labelled support \(\mathcal{S}\)};
\sscansupportsamples{support}

\node[sscan query, right=6mm of support] (query) {};
\node[font=\footnotesize\bfseries, anchor=north] at
  ($(query.north)+(0,-1.5mm)$) {Unlabelled query \(\mathcal{Q}\)};
\sscanquerysamples{query}

\node[sscan panel, fit=(support)(query),
  label={[font=\footnotesize\bfseries]above:
    {Task: support and query samples}}] (taskpanel) {};

\node[sscan encoder, below=5mm of taskpanel] (encoder)
  {\textbf{Encoder per modality and CrossMod Fusion}\\[3mm]
   \([f^{EDA}_\theta(x), f^{ECG}_\phi(x)]\rightarrow F(x)\)};

\draw[sscan flow] (taskpanel.south) -- node[right, sscan note]
  {} (encoder.north);

\node[sscan map, below left=7mm and -20mm of encoder] (supportmaps)
  {Support feature maps\\[3mm]\(\mathcal{F}^s=\{F(x): x\in\mathcal{S}\}\)};
\node[sscan map, below right=7mm and -20mm of encoder] (querymaps)
  {Query feature maps\\[3mm]\(\mathcal{F}^q=\{F(x): x\in\mathcal{Q}\}\)};

\draw[sscan flow] (encoder.south) -- ++(0,-3mm) -| (supportmaps.north);
\draw[sscan flow] (encoder.south) -- ++(0,-3mm) -| (querymaps.north);

\node[sscan prototype, below=5mm of supportmaps] (prototypes)
  {Class prototype maps\\
   $\begin{aligned}
   \\
   \mathbf{P} &= \{\mathbf{p}_c: c \in C\} \\
            \mathbf{p}_c &=
\frac{1}{|\mathcal{F}_c|}
\sum_{F^s_i\in \mathcal{F}_c}
f_\theta(x_i),
\\
\mathcal{F}^s_c &= \{(F^s_i,y_i)\in\mathcal{F}: y_i=c\}
   \end{aligned}$
};
\draw[sscan flow] (supportmaps.south) -- node[right, sscan note]
  {} (prototypes.north);

\node[sscan can, below=5mm of querymaps] (can)
  {Cross Alignment Network\\[1mm]
   \(\displaystyle
     s_{q,c} = \operatorname{CAN}_{\varphi}(\mathbf{p}_c,\mathcal{F}^q)\)};

\draw[sscan flow] ([yshift=26pt]prototypes.east) -- (can.west);
\draw[sscan flow] (querymaps.south) -- (can.north);

\node[sscan output, below=4mm of can] (prediction)
  {Prediction\\
   \(\widehat y_q\sim\mathrm{soft}\max s_{q,c}\)};

\draw[sscan flow] (can.south) -- (prediction.north);

\coordinate (legendstart) at ($(query.east)+(10mm,5mm)$);
\node[sscan sample zero] (legendzero) at (legendstart) {};
\node[font=\scriptsize, right=1.2mm of legendzero] {Class 0};
\node[sscan sample one, below=4mm of legendzero] (legendone) {};
\node[font=\scriptsize, right=1.2mm of legendone] {Class 1};

\end{tikzpicture}

\endgroup

    \caption{Our approach for Subject-As-A-Task Few-Shot Pain Assessment. First, we sample a task,
        which is then passed into an encoder for each modality. These streams are combined using a modified Cross Modality Transformer as introduced by Farmani et al.~\cite{CrossModTransformer2025}, as described in Section~\ref{subsec:crossmod_fusion}.
    The resulting support feature maps are used to compute a prototype per class against which we classify the query
        feature maps using the Cross Alignment Network described in Section~\ref{subsec:can_scoring}.}
    \label{fig:fewshot}
\end{figure}
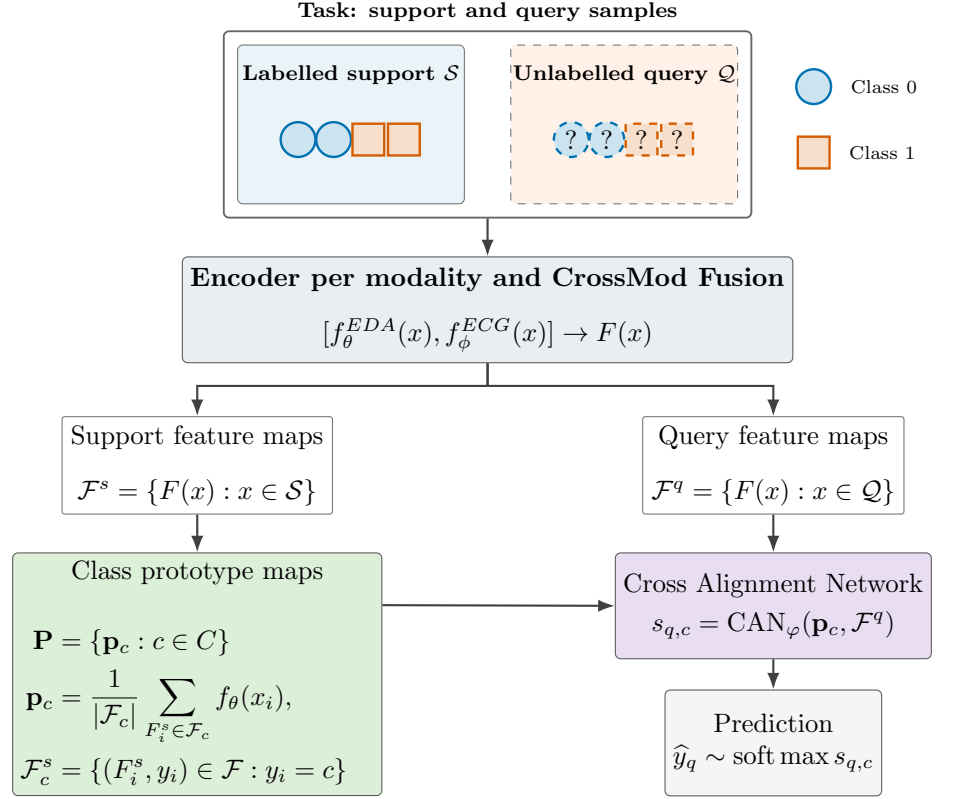

Few-shot learning~\cite{wang2020generalizing} is an instance of meta-learning and describes a setting in which a model
is expected to adapt to a new task, class, or domain using only a small number of labelled examples. Conceptually, the
model first learns generalisable representations from a larger set of source data and then uses a limited support set
to adjust its predictions for a new target case.

Formally, Few-Shot learning considers a distribution over tasks $\mathcal{T} \sim p(\mathcal{T})$, where each task
consists of a small labelled support set $\mathcal{S}$ and a query set $\mathcal{Q}$. In an $N$-way $K$-shot
classification setting with class set \(\mathcal{C}\) and \(|\mathcal{C}|=N\), the support set contains $K$ labelled
examples for each class, i.e., we have
\begin{equation}
\mathcal{S} = \{(x_i, y_i)\}_{i=1}^{N_K},
\qquad
\mathcal{Q} = \{(x_j, y_j)\}_{j=1}^{N_Q}.
\end{equation}
The task is then to use the support samples to condition a model such that it can classify the query samples. In a special case where $k=0$ known as Zero-Shot learning, the model must do so without any conditioning samples. Prototype-based Few-Shot learning solves this by representing each input $x$ through a learned embedding function $f_\theta(x)$ with parameters $\theta$ and computes one
prototype per class as an aggregated embedding of the corresponding support examples, as the mean of the support embeddings as
\begin{equation}
\mathbf{p}_c =
\frac{1}{|\mathcal{S}_c|}
\sum_{(x_i,y_i)\in \mathcal{S}_c}
f_\theta(x_i),
\qquad
\mathcal{S}_c = \{(x_i,y_i)\in\mathcal{S}: y_i=c\},
\end{equation}
where $\mathbf{p}_c$ is the prototype for class $c$ as defined by the labels $y_i$, $\mathcal{S}_c$ is the support set for class $c$, and $f_\theta(x_i)$ is an embedding model that takes an input signal $x_i$ and transforms it to a low dimensional embedding.
A query example $x_j$ is then classified using a $k$-nearest-neighbours scheme based on its distance to the class prototypes by using a softmax over a similarity measure between the embedding of $x_j$ and the class prototypes $\mathbf{p}_c$ as
\begin{equation}
p_\theta(y=c \mid x_j, \mathcal{S})
=
\frac{
\exp\left(d(f_\theta(x_j), \mathbf{p}_c)\right)
}{
\sum_{c'\in\mathcal{C}}
\exp\left(d(f_\theta(x_j), \mathbf{p}_{c'})\right)
},
\end{equation}
where $p_\theta$ is the conditional class probability given the input $x$ and a set of support samples $\mathcal{S}$, and $d$ is a function that computes the similarity between two embeddings. Here, we used the cosine similarity defined as 

\begin{equation}
d(f_\theta(x), \mathbf{p})
=
\frac{f_\theta(x)^{\top}\mathbf{p}}
{\lVert f_\theta(x)\rVert_2 \, \lVert\mathbf{p}\rVert_2}
=
\frac{\sum_{i=1}^{n} f_\theta(x)_i \mathbf{p}_i}
{\sqrt{\sum_{i=1}^{n} f_\theta(x){_i}^2}\sqrt{\sum_{i=1}^{n} \mathbf{p}_i^2}}.
\end{equation}

The embedding parameters $\theta$ are learned by minimising the query-set classification loss across tasks, where we use the support set to anchor the prediction as
\begin{equation}
\mathcal{L}
=
\mathbb{E}_{\mathcal{T}\sim p(\mathcal{T})}
\left[
-\sum_{(x_j,y_j)\in \mathcal{Q}}
\log p_\theta(y_j \mid x_j, \mathcal{S})
\right],
\end{equation}
where $\mathcal{T}$ are task sets consisting of a query set $\mathcal{Q}$ and a support set $\mathcal{S}$.

Thus, learning is not optimised directly for a fixed set of classes, but for the ability to form discriminative class prototypes from only a few labelled support examples and generalise to unseen query examples. Figure~\ref{fig:fewshot} visualises the described approach.

In the LOSO configuration, the held-out subject is not used to construct tasks during training and only serves as the task set during evaluation. We implement two evaluation schemes: Zero-Shot refers to using a prototype bank constructed from encoding all training subjects, while $k$-Shot refers to using a subset of the held-out subjects' data to calibrate the model -- see Subsection~\ref{sec:prototype_bank} for a more detailed explanation.

\subsection{CrossMod Feature Fusion}
\label{subsec:crossmod_fusion}
Our approach implements a multimodal learning scheme, where we leverage two bio-physical signal streams. To combine them, we implement a modified version of the CrossMod approach introduced by Farmani et al.~\cite{CrossModTransformer2025} as a feature-map encoder for EDA and ECG signals.
For each signal window of length $T$ given as $\left[x_F^{\mathrm{EDA}},x_F^{\mathrm{ECG}}\right] \in \mathbb{R}^{T\times 2}$
the two modalities are first processed by independent encoders, resulting in the
modality-specific feature maps $F_i^m = f_\Theta^{m}(x_i)\in \mathbb{R}^{T'\times D_f},$ where $m\in\{\mathrm{EDA},\mathrm{ECG}\}$ is the modality, and $\Theta\in\{\theta,\phi\}$ are the separate parameters of the encoders, and $T',D_f$ are the dimensions of the feature maps. The two branches use the same configuration and therefore preserve a common temporal length $T'$. We process the resulting maps with separate pre-normalised
Transformer encoder stacks, yielding $Z^{\mathrm{EDA}}$ and $Z^{\mathrm{ECG}}$. This allows temporal dependencies
to be modelled within each modality before information is exchanged between the two processing paths.

The fusion stage applies bidirectional multi-head cross-attention \cite{li2021diversity}.Then the two directed feature maps are computed as
\begin{align}
    C^{\mathrm{EDA}\rightarrow\mathrm{ECG}}
    &=
    \operatorname{MHA}
    \left(
    Q=Z^{\mathrm{EDA}},
    K=Z^{\mathrm{ECG}},
    V=Z^{\mathrm{ECG}}
    \right),\\
    C^{\mathrm{ECG}\rightarrow\mathrm{EDA}}
    &=
    \operatorname{MHA}
    \left(
    Q=Z^{\mathrm{ECG}},
    K=Z^{\mathrm{EDA}},
    V=Z^{\mathrm{EDA}}
    \right),
\end{align}
where $MHA$ is a multi-head attention and $Q$, $K$, $V$ are the query, key, and value inputs, respectively, producing the cross-attended map $C^{\mathrm{EDA}\rightarrow\mathrm{ECG}}$ between EDA and ECG, and $C^{\mathrm{ECG}\rightarrow\mathrm{EDA}}$ between ECG and EDA.
Finally, the two cross-attended maps are concatenated along the feature dimension to obtain the single temporal feature
map $F$ as
\begin{equation}
    F
    =
    \left[
    C^{\mathrm{EDA}\rightarrow\mathrm{ECG}},
    C^{\mathrm{ECG}\rightarrow\mathrm{EDA}}
    \right]
    \in
    \mathbb{R}^{T'\times 2D_f}.
    \label{eq:crossmod-feature-map}
\end{equation}

Thus, in the subsequent notation, $T=T'$ and $D=2D_f$. No separate classifier is attached to this representation.
The same encoder produces the support maps, which are averaged class-wise into $P_c$, which are then passed to the Cross Alignment Network described in
Section~\ref{subsec:can_scoring}.

\subsection{Cross-Attention Network Scoring}
\label{subsec:can_scoring}

To efficiently score the alignment of query feature maps with the prototypes and thus be able to classify them, we use the Cross Alignment Network (CAN), as originally proposed by Hou et al.~\cite{hou2019cross} and recently applied to automated pain assessment by
Gkikas et al.~\cite{gkikas2025efficient} and Li et al.~\cite{li2024automatic}. To this end, we form a descriptor of a prototype-query pair. Let 

\begin{align}
    \mu_\mathbf{p}&=\frac{1}{T_\mathbf{p}}\sum_i \mathbf{p}_{d,c} \hspace{1em}\mathrm{and} \\
    \mu_F&=\frac{1}{T_F}\sum_j F_{j}
\end{align}

denote the un-normalised temporal means of the prototypes $\mathbf{p}$ of class $c$, and of the query feature maps $F$. Their concatenation as

\begin{equation}
    d^{\mathrm{pair}}_{\mathbf{p},F}
    =
    [\mu_\mathbf{p},\mu_F,|\mu_\mathbf{p}-\mu_F|,\mu_\mathbf{p}\odot\mu_F],
    \label{eq:can-pair-descriptor}
\end{equation}

amended with an absolute difference, and an element-wise product defines the pair descriptor. A meta-kernel network maps this descriptor to normalised kernels over the opposite temporal axes, as

\begin{align}
    f_{\mathbf{p},F}
    &=
    \operatorname{ReLU}
    \left(d^{\mathrm{pair}}_{\mathbf{p},F}\,W_\mathrm{pair}+b_\mathrm{pair}\right),
    \label{eq:can-meta-hidden}\\
    k_{\mathbf{p}}&=  \operatorname{softmax}\left(f_{\mathbf{p},F}\,W_{\mathbf{p}}+b_{\mathbf{p}}\right),\\    
    k_F&=
    \operatorname{softmax}\left(f_{\mathbf{p},F}\,W_F+b_F\right),
    \label{eq:can-meta-kernels}
\end{align}

where \(f_{\mathbf{p},F}\) is the non-linear transformation of the pair-descriptor \(d^{\mathrm{pair}}_{\mathbf{p},F}\), \(k_{\mathbf{p}}\in\mathbb{R}^{T_F}\) weights query positions when computing prototype-side evidence, whereas \(k_F\in\mathbb{R}^{T_{\mathbf{p}}}\) weights prototype positions when computing query-side evidence, and $W$ and $b$ denote the respective weights and biases of these transformations. Next, we aggregate the attention maps and scale them as

\begin{align}
    u^{\mathbf{p}}
    &=
    \sum_i R_{i,j}\,k_{\mathbf{p},i},
    &
    A_{\mathbf{p}}
    &=
    \operatorname{softmax}\left(u^{\mathbf{p}}/\tau\right),
    \label{eq:can-prototype-attention}\\
    u^F
    &=
    \sum_{j} R_{i,j}\,k_{F,j},
    &
    A_F
    &=
    \operatorname{softmax}\left(u^F/\tau\right),
    \label{eq:can-query-attention}
\end{align}

where \(\tau>0\) controls the concentration of both temporal attention
distributions. The softmax in \(A_{\mathbf{p}}\) operates over \(p\), and the softmax in \(A_F\) operates over \(r\). \(R_{i,j}\) is the temporal correlation matrix, making $u^\mathbf{p}$ and $u^F$ marginals over the correlation matrix in their respective dimension. In the following, we explain how we compute $R_{i,j}$ and how we interpret it in the context of CAN.

First, we normalise every temporal feature vector in the prototype map $\mathbf{p}$ and in the query map $F$ along the respective temportal positions $i$ and $j$ as
\begin{equation}
    \bar{\mathbf{p}}_{i}
    =
    \frac{\mathbf{p}_{i}}{\lVert \mathbf{p}_{i}\rVert_2},
    \qquad
    \bar F_{j}
    =
    \frac{F_{j}}{\lVert F_{j}\rVert_2},
    \label{eq:can-channel-normalisation}
\end{equation}
and construct the temporal correlation matrix as
\begin{equation}
    R_{i,j}
    =
    \left\langle \bar{\mathbf{p}}_{i},\bar F_{j}\right\rangle
    \in\mathbb{R}^{T_{\mathbf{p}}\times T_F},
    \label{eq:can-temporal-correlation}
\end{equation}
where $T_{\mathbf{p}}$ and $T_F$ are the temporal dimensions of the prototype map and the query map, respectively. Thus, each entry measures the cosine similarity between one prototype
position \(d\) and one query position \(r\). This is the temporal analogue of the local correlation layer in the original image-based CAN~\cite{hou2019cross}.

Next, we compute an attention-weighted temporal summary for each branch as
\begin{equation}
    z_{\mathbf{p}}
    =
    \sum_i A_\textbf{p}\,\mathbf{p}_{i},
    \qquad
    z_{F}
    =
    \sum_j A_{F,j}\,F_{j}.
    \label{eq:can-attention-weighted-summaries}
\end{equation}
To retain a stable global representation while allowing pair-conditioned
attention to emphasise informative temporal positions, a learned scalar gate
interpolates between each unweighted temporal mean and its attention-weighted
summary. Specifically, with \(g=\sigma(\gamma)\in(0,1)\), where \(\gamma\) is a
trainable parameter shared by the prototype and query branches, the final
descriptors are
\begin{equation}
    v_\mathbf{p} = (1-g)\mu_\mathbf{p}+g\,z_\mathbf{p},
    \qquad
    v_{F} = (1-g)\mu_F+g\,z_F.
    \label{eq:can-attended-pooling}
\end{equation}
The CAN score is then the cosine similarity between these descriptors for prototype $\mathbf{p}$ and feature map $F$, given as
\begin{equation}
    \rho_{\mathbf{p}, F} =
        \left\langle
        \frac{v_{\mathbf{p}}}{\lVert v_{\mathbf{p}}\rVert_2},
        \frac{v_{F}}{\lVert v_{F}\rVert_2}
    \right\rangle.
    \label{eq:can-detail-score}
\end{equation}
Repeating this computation for all classes yields one similarity score per
query--class pair, which can then be used to compute the class-level similarity score as
\begin{equation}
    s_{\mathbf{p},F}
    =
    \log
    \left(
    \frac{1}{M_{\mathrm{slot}}}
    \sum_{m=1}^{M_{\mathrm{slot}}}
    \exp(\rho_{\mathbf{p}_m,F})
    \right),
    \label{eq:slot-aggregation}
\end{equation}
where $\mathbf{p}_m$ are the $M$ prototypes and $F$ is the feature maps of the query samples computed as described earlier.
\subsection{Zero-Shot Classification via a Prototype Bank}
\label{sec:prototype_bank}
\begin{figure}
    \centering
    \resizebox{\textwidth}{!}{
\begingroup

\tikzset{
  sscan set/.style={
    draw=black!55,
    rounded corners=2pt,
    inner sep=3pt,
    align=center
  },
  sscan query/.style={
    sscan set,
    fill=sscanqueryfill,
    dashed,
    minimum width=52mm,
    minimum height=24mm
  },
  sscan support/.style={
    sscan set,
    fill=sscansupportfill,
    minimum width=54mm,
    minimum height=28mm
  },
  sscan process/.style={
    draw=black!65,
    rounded corners=2pt,
    minimum width=50mm,
    minimum height=14mm,
    inner sep=3pt,
    align=center
  },
  sscan encoder/.style={
    sscan process,
    fill=sscanencoderfill
  },
  sscan map/.style={
    draw=black!55,
    rounded corners=2pt,
    minimum width=46mm,
    minimum height=12mm,
    fill=white,
    align=center,
    inner sep=3pt
  },
  sscan prototype/.style={
    draw=black!65,
    rounded corners=2pt,
    fill=sscanprototypefill,
    minimum width=74mm,
    minimum height=60mm,
    inner sep=3pt,
    align=center
  },
  sscan can/.style={
    sscan process,
    fill=sscancanfill,
    minimum width=56mm,
    minimum height=16mm
  },
  sscan aggregate/.style={
    draw=black!65,
    rounded corners=2pt,
    fill=sscannormfill,
    minimum width=54mm,
    minimum height=23mm,
    inner sep=3pt,
    align=center
  },
  sscan output/.style={
    sscan process,
    fill=black!4,
    minimum width=50mm
  },
  sscan slot/.style={
    draw=black!55,
    rounded corners=1.5pt,
    fill=white!70,
    minimum width=10mm,
    minimum height=7mm,
    inner sep=1pt,
    font=\scriptsize,
    align=center
  },
  sscan flow/.style={
    -{Latex[length=2.2mm]},
    line width=0.8pt,
    draw=black!75
  },
  sscan note/.style={
    font=\scriptsize,
    align=center,
    text=black!70
  },
  sscan sample zero/.style={
    circle,
    draw=sscanclasszero,
    fill=sscanclasszero!20,
    line width=0.9pt,
    minimum size=4.6mm,
    inner sep=0pt
  },
  sscan sample one/.style={
    rectangle,
    draw=sscanclassone,
    fill=sscanclassone!20,
    line width=0.9pt,
    minimum size=4.2mm,
    inner sep=0pt
  },
  sscan query zero/.style={
    sscan sample zero,
    dashed
  },
  sscan query one/.style={
    sscan sample one,
    dashed
  },
    sscan separator/.style={
    draw=black!35,
    dashed,
    dash pattern=on 3pt off 2.5pt,
    line width=0.65pt
  },
}

\newcommand{\sscanquerysamples}[1]{%
  \node[sscan query zero] at ($(#1.center)+(-18mm,-2mm)$) {?};
  \node[sscan query zero] at ($(#1.center)+(-10mm,-2mm)$) {?};
  \node[sscan query zero] at ($(#1.center)+(-2mm,-2mm)$) {?};
  \node[font=\footnotesize] at ($(#1.center)+(6mm,-2mm)$) {\(\cdots\)};
  \node[sscan query one] at ($(#1.center)+(14mm,-2mm)$) {?};
  \node[sscan query one] at ($(#1.center)+(22mm,-2mm)$) {?};
}

\newcommand{\sscanksupportsamples}[1]{%
  \node[font=\scriptsize,anchor=west]
    at ($(#1.center)+(-24mm,4mm)$) {Class \(c\)};
  \foreach \x in {-6,2,10} {
    \node[sscan sample zero] at ($(#1.center)+(\x mm,4mm)$) {};
  }
  \node[font=\footnotesize]
    at ($(#1.center)+(19mm,4mm)$) {\(\cdots\)};

  \node[font=\scriptsize,anchor=west]
    at ($(#1.center)+(-24mm,-8mm)$) {Class \(c'\)};
  \foreach \x in {-6,2,10} {
    \node[sscan sample one] at ($(#1.center)+(\x mm,-8mm)$) {};
  }
  \node[font=\footnotesize]
    at ($(#1.center)+(19mm,-8mm)$) {\(\cdots\)};

  \node[font=\scriptsize,anchor=south east]
    at ($(#1.south east)+(-2mm,-1mm)$) {\(k\) samples / class};
}

\resizebox{\linewidth}{!}{%
\begin{tikzpicture}[x=1mm,y=1mm]

\coordinate (leftcolumn)   at (-68,0);
\coordinate (middlecolumn) at (  0,0);
\coordinate (rightcolumn)  at ( 68,0);

\node[font=\bfseries, align=center]
  at ($(leftcolumn)+(0,0)$) {Zero-Shot};

\node[font=\bfseries, align=center]
  at ($(middlecolumn)+(0,0)$) {Shared query pipeline};

\node[font=\bfseries, align=center]
  at ($(rightcolumn)+(0,0)$) {\(k\)-shot};
\draw[sscan separator]
  (-35,7) -- (-35,-158);

\draw[sscan separator]
  (35,7) -- (35,-158);
\node[sscan query] (query)
  at ($(middlecolumn)+(0,-18)$) {};

\node[font=\footnotesize\bfseries, anchor=north]
  at ($(query.north)+(0,-2mm)$)
  {Unlabelled query \(\mathcal Q\)};

\sscanquerysamples{query}

\node[sscan encoder] (query-encoder)
  at ($(middlecolumn)+(0,-46)$)
  {\textbf{Encoder}\\[1.5mm]
   \(\bigl[f^{\mathrm{EDA}}_{\psi}(x),f^{\mathrm{ECG}}_{\phi}(x)\bigr]
   \longrightarrow F(x)\)};

\draw[sscan flow]
  (query.south) -- (query-encoder.north);

\node[sscan map] (querymaps)
  at ($(middlecolumn)+(0,-70)$)
  {\textbf{Query feature maps}\\[1.5mm]
   \(\mathcal F^q=\{F(x):x\in\mathcal Q\}\)};

\draw[sscan flow]
  (query-encoder.south) -- (querymaps.north);

\node[sscan can] (can)
  at ($(middlecolumn)+(0,-96)$)
  {\textbf{Cross Alignment Network}\\[1.5mm]
   \(\rho_{q,c,m}=
   \operatorname{CAN}_{\varphi}\!\bigl(P_{c,m},\mathcal F^q\bigr)\)};

\draw[sscan flow]
  (querymaps.south) -- (can.north);

\node[sscan aggregate] (aggregation)
  at ($(middlecolumn)+(0,-122)$)
  {\textbf{Slot aggregation}\\[1.5mm]
   \(\displaystyle
   s_{\mathbf p,F}
   =
   \log\!\left(
   \frac{1}{M_{\mathrm{slot}}}
   \sum_{m=1}^{M_{\mathrm{slot}}}
   \exp\!\bigl(\rho_{q,c,m}\bigr)
   \right)\)};

\draw[sscan flow]
  (can.south) -- (aggregation.north);

\node[sscan output] (prediction)
  at ($(middlecolumn)+(0,-147)$)
  {\textbf{Prediction}\\[1.5mm]
   \(\widehat y \sim
   \operatorname{softmax}_{c}\bigl(s_{\mathbf p,F}\bigr)\)};

\draw[sscan flow]
  (aggregation.south) -- (prediction.north);

\node[sscan prototype, minimum height=48mm] (bank)
  at ($(leftcolumn)+(-20,-96)$) {};

\node[font=\footnotesize\bfseries, anchor=north]
  at ($(bank.north)+(0,-2mm)$)
  {Prototype bank};

\node[font=\scriptsize, align=center, anchor=north]
  at ($(bank.north)+(0,-9mm)$)
  {\(\displaystyle
    \Pi \in
    \mathbb{R}^{|\mathcal C|\times M_{\mathrm{slot}}\times T_{\mathbf p}\times D}\)};

\node[font=\scriptsize, anchor=east]
  at ($(bank.center)+(-22mm,1.5mm)$)
  {Class \(c\)};

\node[sscan slot, anchor=east] at ($(bank.center)+(-4mm,1.5mm)$) {1};
\node[sscan slot, anchor=east] at ($(bank.center)+(8mm,1.5mm)$) {2};
\node[font=\scriptsize, anchor=east] at ($(bank.center)+(15mm,1.5mm)$) {\(\cdots\)};
\node[sscan slot, minimum width=12mm, anchor=east]
  at ($(bank.center)+(29mm,1.5mm)$) {\(M\)};

\draw[black!55,dashed,line width=0.6pt]
  ($(bank.center)+(-24mm,-7mm)$) --
  ($(bank.center)+(32mm,-7mm)$);

\node[font=\scriptsize, anchor=east]
  at ($(bank.center)+(-22mm,-15mm)$)
  {Class \(c'\)};

\node[sscan slot, anchor=east] at ($(bank.center)+(-4mm,-15mm)$) {1};
\node[sscan slot, anchor=east] at ($(bank.center)+(8mm,-15mm)$) {2};
\node[font=\scriptsize, anchor=east] at ($(bank.center)+(15mm,-15mm)$) {\(\cdots\)};
\node[sscan slot, minimum width=12mm, anchor=east]
  at ($(bank.center)+(29mm,-15mm)$) {\(M\)};

\node[sscan note, text width=40mm, anchor=south]
  at ($(bank.south)+(0,-1mm)$)
  {precomputed offline};

\draw[sscan flow]
  (bank.east) -- (can.west);

\node[sscan support] (support)
  at ($(rightcolumn)+(0,-18)$) {};

\node[font=\footnotesize\bfseries, anchor=north]
  at ($(support.north)+(0,-2mm)$)
  {Labelled support \(\mathcal S^{(s,k)}\)};

\sscanksupportsamples{support}

\draw[sscan flow]
  (support.south) -- (query-encoder.east);

\node[sscan map] (supportmaps)
  at ($(rightcolumn)+(0,-70)$)
  {\textbf{Support feature maps}\\[1.5mm]
   \(\mathcal F^{(s,k)}=\{F(x):x\in\mathcal S^{(s,k)}\}\)};

\draw[sscan flow]
  (query-encoder.south) -- (supportmaps.north);

\node[sscan prototype, minimum height=25mm] (kproto)
  at ($(rightcolumn)+(20,-106)$) {};

\node[font=\footnotesize\bfseries, anchor=north]
  at ($(kproto.north)+(0,-2mm)$)
  {Subject-specific prototype slots per class};

\node[font=\scriptsize, align=center, anchor=north]
  at ($(kproto.north)+(0,-7mm)$)
  {\(\displaystyle
    \Pi^{(s,k)}_{c,m}
    =
    \operatorname{Init}_{m}\!\left(
      \{F(x^{(s)}_{c,i})\}_{i=1}^{k}
    \right)\)};

\draw[sscan flow]
  (supportmaps.south) -- (kproto.north);

\draw[sscan flow]
  (kproto.west) -- (can.east);

\node[sscan slot, anchor=east] at ($(kproto.center)+(-12mm,-6.5mm)$) {\(P_1\)};
\node[sscan slot, anchor=east] at ($(kproto.center)+(0mm,-6.5mm)$) {\(P_2\)};
\node[font=\scriptsize, anchor=east] at ($(kproto.center)+(6.5mm,-6.5mm)$) {\(\cdots\)};
\node[sscan slot, minimum width=12mm, anchor=east]
  at ($(kproto.center)+(20mm,-6.5mm)$) {\(P_C\)};

\end{tikzpicture}%
}

\endgroup}
    \caption{Our method supports two modes of inference: in the Zero-Shot setting (left), we use a global precomputed bank to initialize the prototypes. Thus, this mode does not require any labelled samples from the subject. In the $k$-Shot case (right), we use a portion of the held-out subjects samples to initialize the prototypes (the support set) and classify the remainder (the query set). Thus, the Zero-Shot setup is the aggregated population level classifier and the $k$-Shot setup is the personalised individual classifier.}
    \label{fig:zero-shot-bank}
\end{figure}
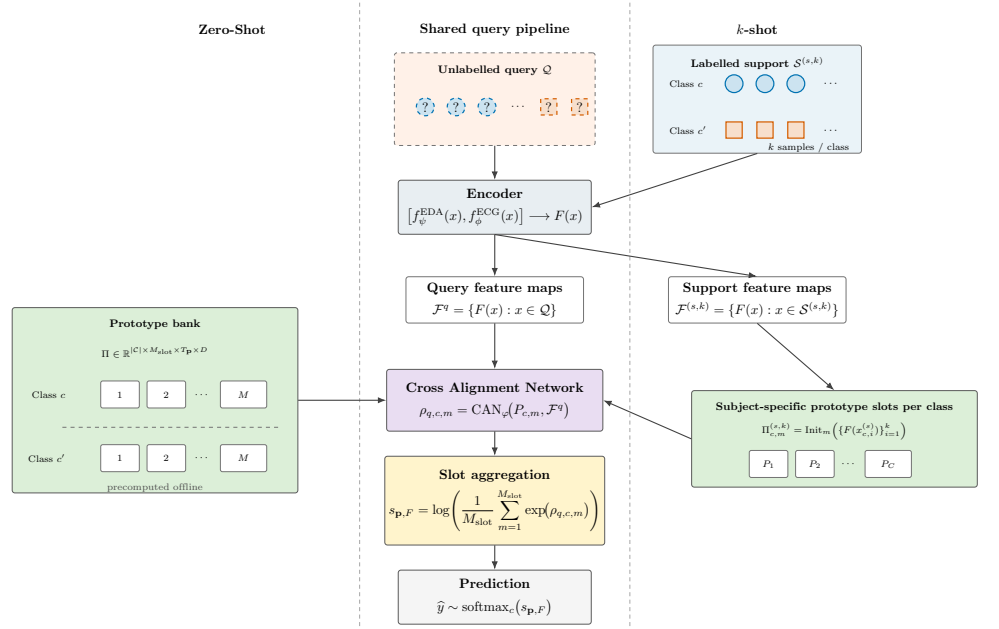
So far, our method is only able to classify samples for the held-out subject if we already have at least one labelled
sample. To enable Zero-Shot classification~\cite{wang2019survey}, i.e., requiring no previous samples, we introduce a
prototype bank that acts as a substitute for the support samples by using persistent temporal prototype slots. Figure~\ref{fig:zero-shot-bank} visualizes the Zero-Shot and $k$-Shot inference modes.

The prototype bank \(\Pi\) contains \(M_{\mathrm{slot}}\) slots per class \(c\),
\begin{equation}
    \Pi
    \in
    \mathbb{R}^{|\mathcal{C}| \times M_{\mathrm{slot}} \times T_{\mathbf{p}} \times D},
    \qquad
    \Pi_{c,m}\in\mathbb{R}^{T_{\mathbf{p}}\times D},
    \label{eq:prototype-bank}
\end{equation}
where \(\mathcal{C}\) is the task class set and \(m\in\{1,\ldots,M_{\mathrm{slot}}\}\) indexes multiple learned
prototypes for the same class. Multiple slots allow the bank to represent within-class heterogeneity instead of forcing
each class into a single temporal template. The memory slots are computed by randomly drawing 258 samples per class (three samples per training subject, i.e., $3\times86=258)$ and
per slot without replacement over all training subjects, encoding them, and then averaging their feature maps. Class-level similarity is obtained with the same log-mean-exp slot aggregation as in the Few-Shot setting, as given by Eq.~\eqref{eq:slot-aggregation}. When we refer to Zero-Shot in our experiments, the evaluation is using the prototype memory as a substitute for the support-based prototypes as described earlier.

\section{Results}
\label{sec:results}
We evaluated our method on the BioVid Heat Pain Database benchmark~\cite{Walter2013BioVid}, on the SenseEmotion Dataset~\cite{velana2016senseemotion}, and on the PainMonit Experimental Dataset (PMED) ~\cite{gouverneur2024experimental}. All experiments were
designed with our three research questions in mind, namely \textit{(i) To what extent can personalisation improve
automated pain assessment under inter-individual variability?}, \textit{(ii) Do the benefits of personalisation vary
across individuals?}, and \textit{(iii) Do the benefits of personalisation vary across pain levels?} We report our results from a single Leave-one-Subject-out (LOSO) run averaged over the metrics computed on the held-out subject. Following Thiam et al.~\cite{Thiam2021DeepPhys}, we used the augmented data for training and the original, unaltered samples for evaluation -- see Figure~\ref{fig:data_augmentation} for more details. Hyperparameters as reported in Table~\ref{tab:hyperparameters} were found by an Optuna~\cite{akiba2019optuna} search on a reduced number of iterations.

\subsection{Personalisation for Automated Pain Assessment}
\label{sec:personalisation_overall}
\begin{table}[ht]
\centering
\caption{Overview of reported pain recognition performance by algorithm/model on the SenseEmotion dataset. All values are reported for Leave-One-Subject-Out (LOSO) cross-validation. Accuracy is reported with standard deviation in parentheses where available. \textbf{Bold} indicates best Zero-Shot method and \underline{underlined} indicates that the personalised $k$-Shot classifiers beats previous methods.}
\label{tab:senseemotion_pain_recognition_performance}
\begin{tabular}{lll}
\toprule
\textbf{Method} & \textbf{Binary} & \textbf{Four Classes} \\
\midrule
Thiam et al.~\cite{thiam2019multi}
& \textbf{83.39\%} 
& \textbf{43.89\%} \\

Thiam et al.~\cite{thiam2021multi}
& 81.05\%
& 40.80\% \\

Thiam et al.~\cite{thiam2025dealing}
& 81.50\%
& n/a \\

Thiam et al.~\cite{thiam2020two}
& 64.35\%
& n/a \\

Kessler et al.~\cite{kessler2017multimodal}
& 71.85\%
& n/a \\
\midrule
Ours (Zero-Shot) & 82.37\% ($\pm 10.9$) & 41.88\% ($\pm 6.6$) \\
Ours ($k$-Shot) & \underline{83.43\%} ($\pm 11.2$) & \underline{44.08\%} ($\pm 9.1$) \\
\bottomrule
\end{tabular}
\end{table}
\begin{table}[ht]
\centering
\caption{Overview of reported pain recognition performance by algorithm/model on the BioVid dataset. All values are reported for Leave-One-Subject-Out (LOSO) cross-validation. Accuracy is reported with standard deviation in parentheses where available. \textbf{Bold} indicates best Zero-Shot method and \underline{underlined} indicates that the personalised $k$-Shot classifiers beats previous methods.}
\label{tab:biovid_pain_recognition_performance}
\begin{tabular}{lll}
\toprule
\textbf{Method} & \textbf{Binary} & \textbf{Five Classes} \\
\midrule
Farmani et al.~\cite{CrossModTransformer2025} & \textbf{87.52\% ($\pm 11.0$)} & n/a \\
Aslam et al.~\cite{aslam2025learning} & 86.90\% & n/a \\
Li et al.~\cite{li2024automatic} & 86.21\% & 38.03\% \\
Lu et al.~\cite{Lu2023PainAttnNet} & 85.56\% & 34.46\% \\
Thiam et al.~\cite{thiam2025dealing} & 85.32\% ($\pm13.7$) & n/a \\
Jiang et al.~\cite{jiang2024personalized} & 84.58\% ($\pm 13.3$)& \textbf{39.24\%} ($\pm 8.65$) \\
Thiam et al.~\cite{thiam2019exploring} & 84.40\% ($\pm 14.4$)& 36.54\% ($\pm 8.55$)\\
Thiam et al.~\cite{Thiam2021DeepPhys} & 84.20\% ($\pm 13.7$) & 35.50\% ($\pm 7.9$)\\
Steuer \& Schwenker~\cite{Steur2025Supervised} & 84.22\% ($\pm 13.2$) & n/a \\
Hihn \& Schwenker~\cite{hihn2026efficient} & 84.00\% ($\pm15.8$) & 36.20\% ($\pm9.7$) \\
\midrule
Ours (Zero-Shot) & 85.75\% ($\pm$ 13.5) & 35.49\% ($\pm 7.00$) \\
Ours ($k$-Shot) & 86.25\% ($\pm$ 14.2) & \underline{40.06\%} ($\pm 12.1$) \\
\bottomrule
\end{tabular}
\end{table}

\begin{table}[ht]
\centering
\caption{Overview of reported pain recognition performance by algorithm/model on the PainMonit dataset. All values are reported for Leave-One-Subject-Out (LOSO) cross-validation. Accuracy is reported with standard deviation in parentheses where available. \textbf{Bold} indicates best Zero-Shot method and \underline{underlined} indicates that the personalised $k$-Shot classifiers beats previous methods.}
\label{tab:painmonit_pain_recognition_performance}
\begin{tabular}{lll}
\toprule
\textbf{Method} & \textbf{Binary} & \\
\midrule
Luebke et al.~\cite{luebke2023objective}
& \textbf{93.62\%}
& \\
Gouverneur et al.~\cite{gouverneur2023explainable}
& 93.26\%
& \\
Gouverneur et al.~\cite{gouverneur2023explainable}
& 89.79\% ($\pm$ 11.68)
& \\
Gouverneur et al.~\cite{gouverneur2021comparison}
& 87.41\% ($\pm 11.99$)
& \\
Gouverneur et al.~\cite{gouverneur2023explainable}
& 91.09\% ($\pm 9.70$)
& \\
\midrule
Ours (Zero-Shot) & 90.47\% ($\pm$ 9.23) & \\
Ours ($k$-Shot) & 91.25\% ($\pm$ 8.91) & \\
\bottomrule
\end{tabular}
\end{table}

To answer the first research question (\textit{To what extent can personalisation improve automated pain assessment under inter-individual variability?}), we provide results from the automated pain assessment benchmark in binary
(no pain vs. high pain stimulus) and full multi-classification settings in Tables~\ref{tab:senseemotion_pain_recognition_performance},~\ref{tab:biovid_pain_recognition_performance}, and~\ref{tab:painmonit_pain_recognition_performance}. To avoid single dataset bias, we evaluate our method on three datasets and in several classification settings. This gives us an overview of our method's performance compared to previously published models.

\begin{figure}[ht]
\centering
\includegraphics[width=0.9\linewidth]{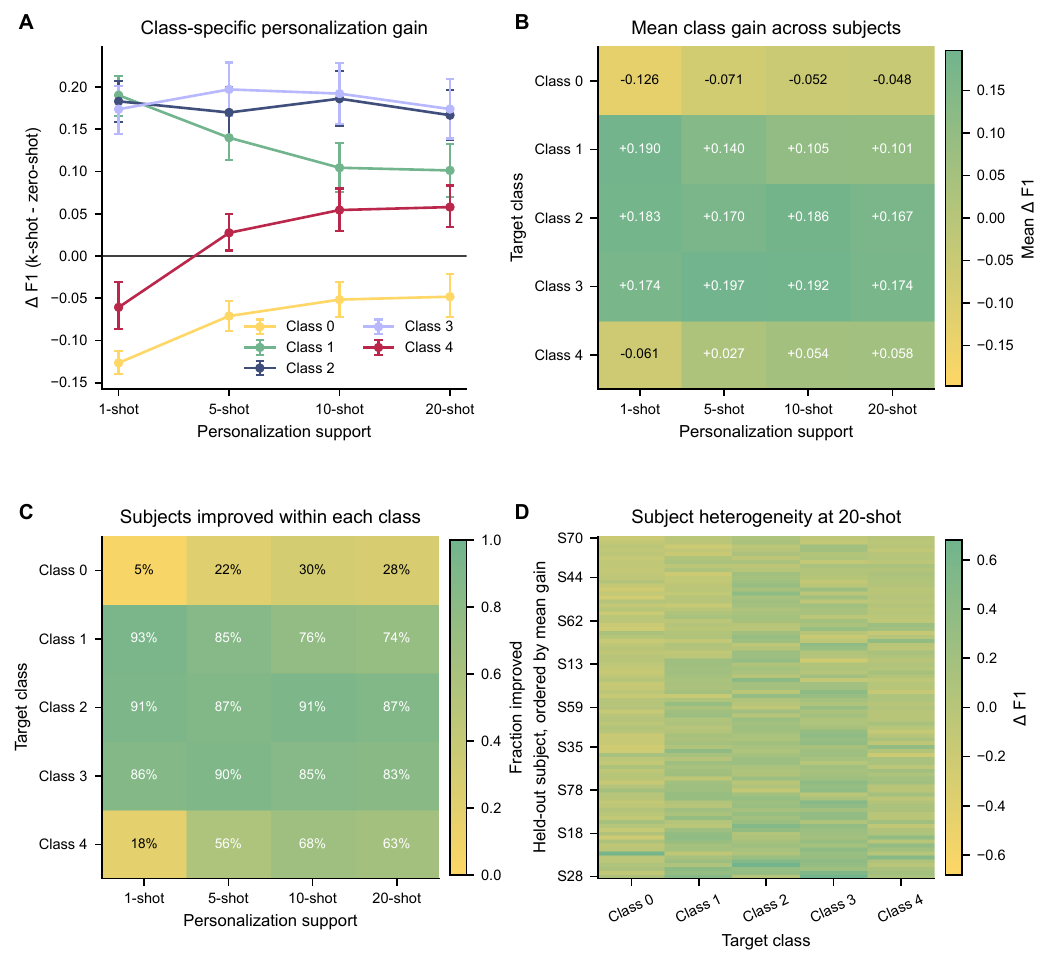}
\caption{Personalisation benefits per class on the BioVid dataset. Panels~A and~B show that the intermediate pain levels 1, 2, and 3 have the largest mean F1 gains from personalisation. Their respective F1 scores improve on average, indicating that personalisation contributes most for pain levels that vary only little in magnitude, compared with the extremes 0 and 4. Panels~C and~D show how subjects benefit on a per-class basis, indicating the same pattern of improvement in classes 1 to 3.}
    \label{fig:benefits_per_class}
\end{figure}

On the BioVid dataset and in the Zero-Shot case, our method reaches $85.75\%$ binary accuracy and $35.49\%$ multi-class accuracy; in the $k$-Shot case,
it reaches $86.25\%$ and $40.06\%$, respectively. On the SenseEmotion dataset, our method reaches $82.37\%$ binary accuracy and $41.88\%$ multi-class accuracy in the Zero-Shot case; in the k-Shot case, it reaches $83.43\%$ and $44.08\%$, respectively. On PMED, our method reaches $90.47\%$ binary accuracy in the Zero-Shot case and $91.25\%$ in the $k$-Shot case; no multi-class benchmark has been reported for this dataset, so we evaluate it in the binary setting only. In the $k$-Shot adaptive case, the method uses labels from the held-out
subject, so this setting is not directly comparable to Zero-Shot methods, which is why we compare methods directly only in the Zero-Shot setting and report $k$-Shot results to demonstrate the potential benefits of personalisation as we have introduced it in our work. The Zero-Shot performance of our method is comparable to previous methods. Under the present episodic evaluation protocol,
$k$-Shot classification produced higher mean multi-class performance than state-of-the-art methods on the BioVid dataset ($40.06\%$ vs. $39.24\%$) and on the SenseEmotion dataset ($44.08\%$ vs. $43.89\%$), suggesting the largest benefit of personalisation is in the complex full-class setting. In the binary setting, our method yields comparable performance. We investigate the personalisation benefits across classes furhter in Section~\ref{sec:personalisation_class} and in Figure~\ref{fig:benefits_per_class}.

In Table~\ref{tab:binary_mc_comparison} we investigate additional binary classification settings. Our method shows benefits in the $k$-Shot setting in all classification problems and yields higher performance in the Zero-Shot setting for $T_0$ vs. $T_1$, $T_0$ vs. $T_2$ on BioVid and $T_0$ vs. $T_2$ on SenseEmotion compared to other methods. The setting $T_0$ vs. $T_1$ poses the most difficult classification problem, as the corresponding pain stimuli are closest and thus the bodily reactions are similar. In that setting, our method yields higher accuracy than previous methods on the BioVid dataset and comparable accuracy on the SenseEmotion dataset. Additionally, all personalised $k$-Shot classifiers outperform previous methods, except for the binary cass in BioVid, where Farmani et al.~\cite{CrossModTransformer2025} report 87.52\%, above our k-Shot 86.25\%.

Overall, the benefits of personalisation as we have formalised it in our study vary between settings but are consistent in that it yields higher accuracy. In all settings and datasets, the increase in accuracy in the $k$-Shot setting indicates the benefit of personalisation, which we investigate further in the next sections.

\begin{table}[ht]
\centering
\caption{Comparison of additional binary pain recognition results. \textbf{Bold} indicates best Zero-Shot method and \underline{underlined} indicates that the personalised $k$-Shot classifiers beats previous methods.}
\label{tab:binary_mc_comparison}
\begin{tabular}{lll}
\toprule
& \textsc{BioVid} &\\
\toprule
\textbf{Method} 
& \textbf{T$_0$ vs. T$_1$} 
& \textbf{T$_0$ vs. T$_2$} \\
\midrule
Thiam et al.~\cite{thiam2019exploring} 
& 61.15\% ($\pm 12.20$) 
& 66.81\% ($\pm 15.92$) \\
Gkikas et al.~\cite{gkikas2021multi}
& 52.38\% & 52.78\% \\
Werner et al.~\cite{werner2014automatic}
& 48.70\% & 51.60\% \\
Gkikas et al.~\cite{gkikas2021multi}
& 62.82\% & 63.68\% \\
Lopez \& Picard~\cite{lopez2018continuous}
& 56.44\% & 59.40\% \\
\midrule
Ours (Zero-Shot)
& \textbf{61.26\%} ($\pm 13.02$) & \textbf{67.07\%} ($\pm 14.91$) \\
Ours ($k$-Shot)
& \underline{62.25\%} ($\pm 14.20$)
& \underline{68.06\%} ($\pm 16.56$)\\
\toprule
\textbf{Method} 
& \textbf{T$_0$ vs. T$_3$} 
& \textbf{T$_1$ vs. T$_4$} \\
\midrule
Thiam et al.~\cite{thiam2019exploring} 
& \textbf{76.29\%} ($\pm 14.62$) 
& \textbf{76.72\%} ($\pm 15.02$) \\
Gkikas et al.~\cite{gkikas2021multi}
& 55.37\% & n/a \\
Werner et al.~\cite{werner2014automatic}
& 56.50\% & n/a \\
Gkikas et al.~\cite{gkikas2021multi}
& 66.12\% & n/a \\
Lopez et al.~\cite{lopez2018continuous}
& 66.00\% & n/a \\
\midrule
Ours (Zero-Shot)
& 76.15\% ($\pm 16.11$)
& 76.67\% ($\pm 15.33$) \\
Ours ($k$-Shot)
& \underline{77.01\%} ($\pm 14.34$)
& \underline{76.98\%} ($\pm 11.788$) \\
\midrule
& \textsc{SenseEmotion} & \\
\toprule
\textbf{Method} 
& \textbf{T$_0$ vs. T$_1$} 
& \textbf{T$_0$ vs. T$_2$} \\
\midrule
Thiam et al.~\cite{thiam2019multi} & \textbf{52.74\%} ($\pm 7.64$) & 60.91\% ($\pm12.69$)\\
\midrule
Ours (Zero-Shot) & 52.46\% & \textbf{62.44\%} ($\pm 9.66$) \\ 
Ours ($k$-Shot) & \underline{53.86\%} & \underline{63.89\%} ($\pm 7.96$)\\
\bottomrule
\end{tabular}
\end{table}

\subsection{Personalisation Benefits per Subject}
\label{sec:personalisation_subject}
\begin{figure}[htbp]
\centering
\includegraphics[width=\linewidth]{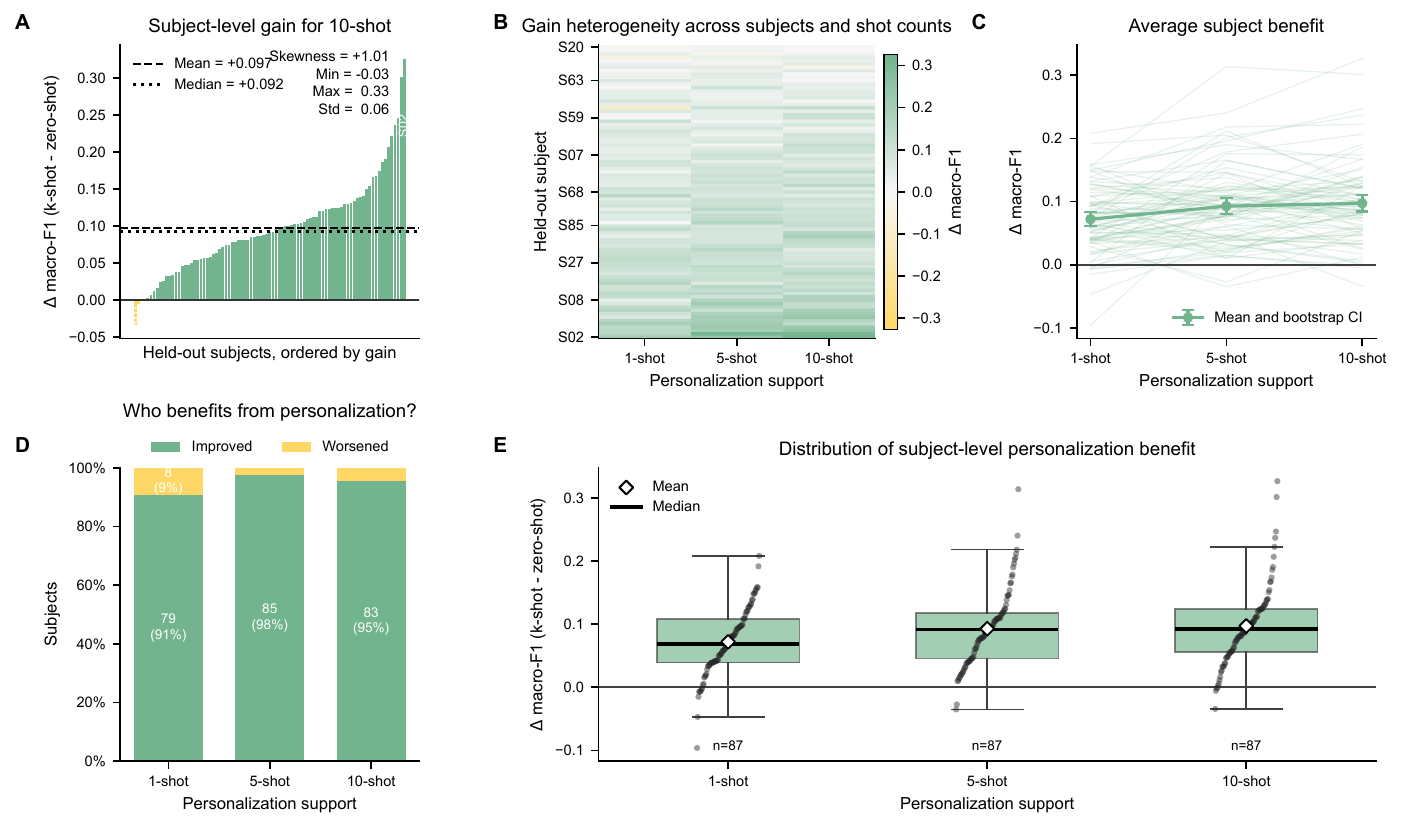}
\caption{Personalisation benefits per subject. A vast majority of 91\% to 98\% of subjects benefit from personalisation in the BioVid full classification setting, i.e., using part of their samples to initialize the prototypes (see Panels A, B, D). Panel A shows a mean improvement in macro F1 is $+0.097$. Panel E shows the distribution of the improvements in the different $k$ settings.}
    \label{fig:personalization_benefit_subject}
\end{figure}
To answer the second research question (\textit{Do the benefits of personalisation vary across individuals?}), we
analyse subject-level changes in classification performance on the BioVid dataset in Figure~\ref{fig:personalization_benefit_subject}. The results in the 10-shot setting show
that personalisation does not affect all held-out subjects equally: although almost all subjects improve, some experience lower macro-F1 in their respective LOSO runs, ranging from $-0.03$ to $0.33$ with a mean of $0.097$, median of $0.092$, and a standard deviation of $0.06$, as shown in Panel A of Figure~\ref{fig:personalization_benefit_subject}. Panel D in Figure~\ref{fig:personalization_benefit_subject} further support the conclusion that 91\% to 98\% of subjects benefit from personalisation.

However, the improvement varies with the number of support samples available. At $k=1$, each class contributes only one labelled sample to the support statistics. Noisy or atypical support windows can therefore
produce unrepresentative prototypes, leading to the lowest benefit. Once we add five or more samples, the fraction of subjects that benefits increases, as other, more representative samples may counterbalance noisy samples.

The evaluation protocol introduces an additional source of variation. The Zero-Shot condition uses all held-out query
samples once, whereas the $k$-Shot condition aggregates 500 sampled support/query episodes. It therefore estimates
expected support-conditioned episodic performance rather than performance on the same fixed query set before and after
adding support. Unfavourable support or query samples can lower a subject's estimated $k$-Shot performance even when a
different support set would be beneficial. 

In summary, personalisation improves mean macro-F1, but its benefit varies substantially across subjects. Persistent 
gain patterns and large random subject effects provide evidence of stable subject-level improvements, as supported by the analyses in Figure~\ref{fig:personalization_benefit_subject}.

\begin{figure}[ht]
    \centering
    \includegraphics[width=0.99\linewidth]{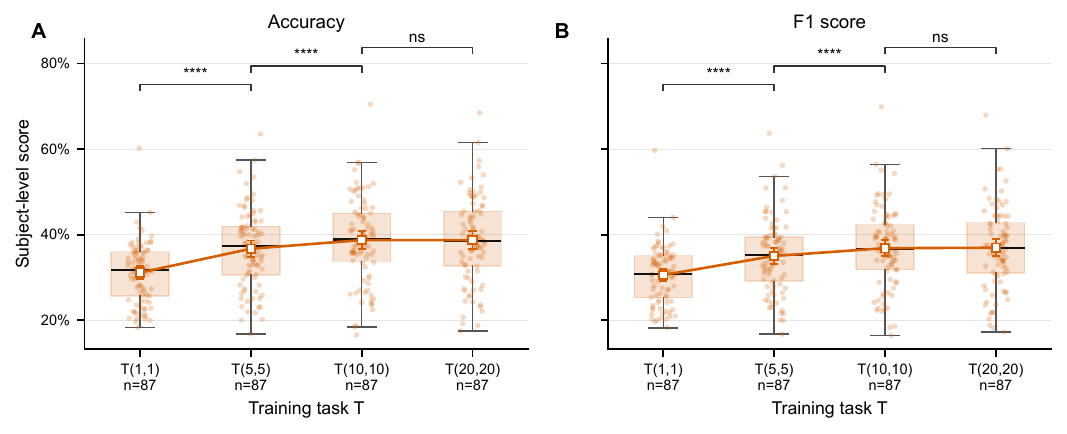}
    \includegraphics[width=0.99\linewidth]{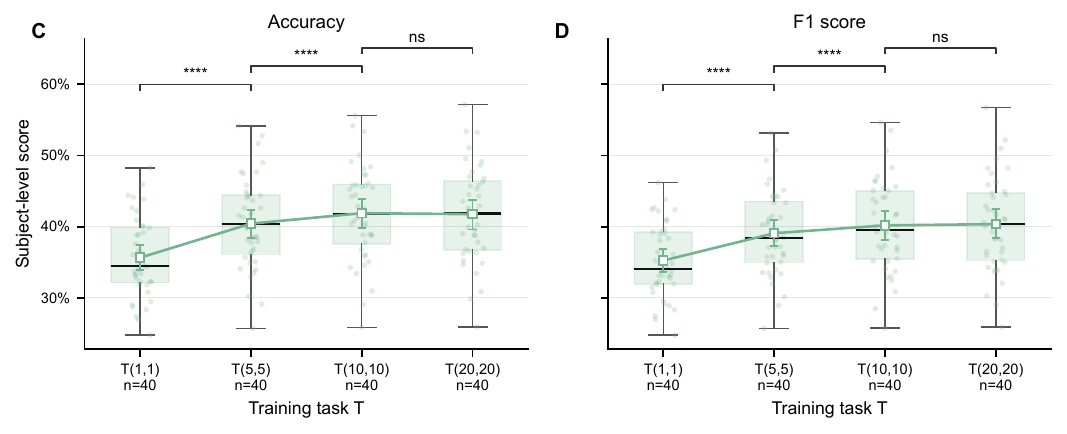}
    \caption{Different settings of training for the BioVid (A+B) and the SenseEmotion (C+D) dataset. In all settings increasing the training task size ($k,q$ values) improves the results with a ceiling on $k,q=10$ where we do not observe significantly higher results. All significant $p$-values are $< 0.001$.}
    \label{fig:kq_setting_fold_metrics}
\end{figure}

\subsection{Personalisation Benefits per Class}
\label{sec:personalisation_class}
To answer the third research question (\textit{Do the benefits of personalisation vary across pain levels?}), we
analysed the average gain in the F1 score on a per-class basis in Figure~\ref{fig:benefits_per_class}. Classes 1, 2, and 3
have mean F1 gains of $0.190$, $0.183$, and $0.174$, respectively, whereas the baseline class 0 and the maximum pain level
4 have gains of $-0.126$ and $0.061$, respectively. This class-wise pattern is consistent with inter-subject variation being concentrated in the intermediate pain levels rather than at the two extremes. The negative change in classes 0 and 4 indicates that the model is moving some of its descriptive power from the extremes to the intermediate classes, facilitating a trade-off in which class-0 F1 decreases while intermediate-class F1 increases.

In summary, the multi-class analyses show a consistent redistribution of performance across pain
levels. Personalisation improves the hard intermediate classes but reduces F1 for class 0 across the inspected support sizes and, at small $k$, reduces F1 for the highest pain class. 

\subsection{Ablation Experiments}
\label{sec:ablation}

\begin{figure*}[ht]
      \centering
      \includegraphics[width=0.9\textwidth]
          {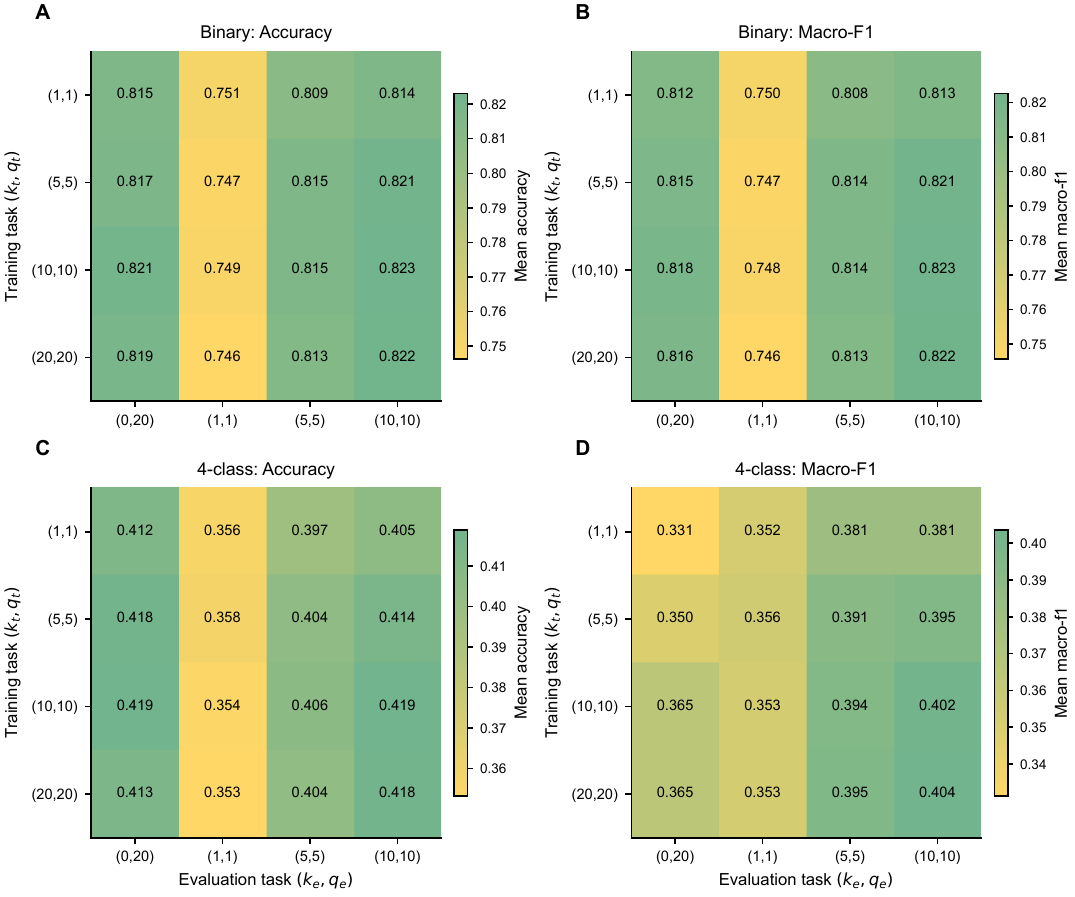}
      \caption{Mean held-out performance across training and evaluation
      conditions for binary classification (top) and multi-class classification
      (bottom) on the SenseEmotion dataset. Columns show accuracy and macro-F1. The Zero-Shot condition
      $(k_e,q_e)=(0,20)$ uses learned prototype memory, whereas the remaining
      conditions use target-subject support.}
      \label{fig:train-eval-heatmaps}
  \end{figure*}

We have designed a set of ablation experiments to verify the stability and performance of our proposed method. We investigate the choice of signals used, and the number of support samples during evaluation and during training.

\subsubsection{Training and Evaluation Task Setup}
\label{sec:training_eval_ablation}
\begin{figure}
    \centering
    \includegraphics[width=0.95\linewidth]{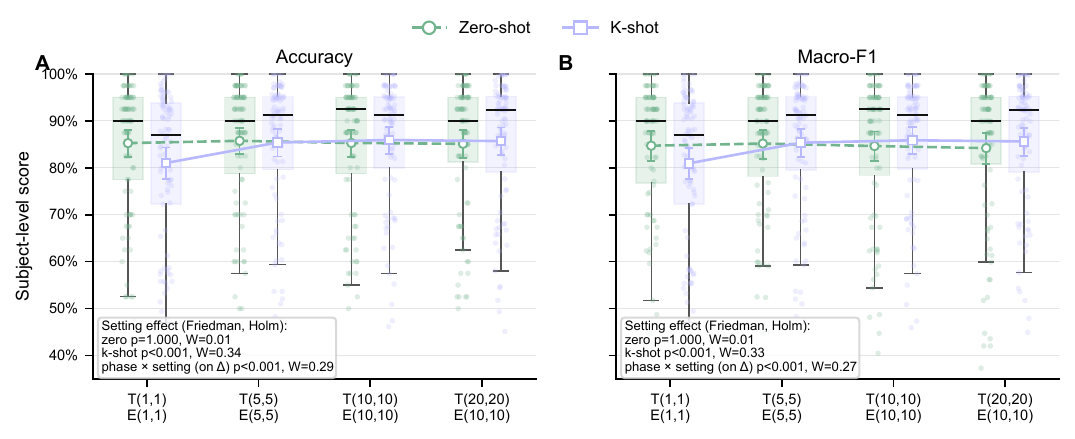}
    \includegraphics[width=0.95\linewidth]{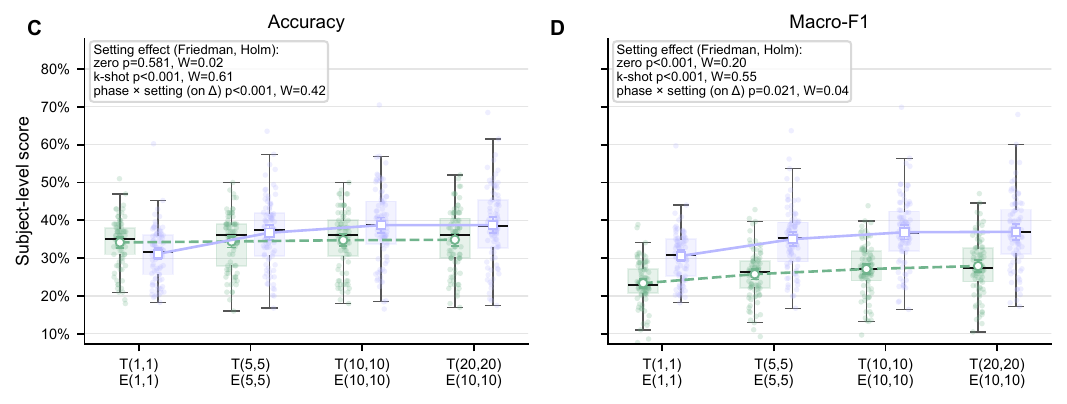}
\caption{Effect of training and evaluation task size on BioVid binary (A+B) and five-class (C+D) performance. Points represent held-out subjects, and boxes show the score distributions. Here, T and E denote the training and configured $k$-shot evaluation tasks, respectively.}
    \label{fig:zero_kshot_by_setting}
\end{figure}

In Figure~\ref{fig:zero_kshot_by_setting} we evaluate the number of support and query samples during training, indicated as $k_t$ and $q_t$, respectively in the personalised $k$-shot setting on the BioVid (\ref{fig:zero_kshot_by_setting}.A+\ref{fig:zero_kshot_by_setting}.B) and one the SenseEmotion (\ref{fig:zero_kshot_by_setting}.C+\ref{fig:zero_kshot_by_setting}.D) dataset. The mean accuracy and F1 score improve significantly up to $k_t,q_t=5$, and the gains are smaller beyond that point. These results suggest that increasing the number of
support samples during training helps the model extract more robust feature spaces, but the effect hits a ceiling after 10 samples.

Figure~\ref{fig:train-eval-heatmaps} shows results of an ablation experiment on the SenseEmotion dataset that inspects the influence of the number of support samples used during evaluation. As expected, increasing the number of support samples
improves average accuracy, but the effect occurs only on the personalised $k$-Shot settings -- we will discuss this further in Figure~\ref{fig:zero_kshot_by_setting}. These results
indicate that an increased number of support samples during the evaluation enables the model to find prototypes that are
less noisy and more representative.

Additionally, we examined whether the support/query configuration used during training affected zero-shot and subject-adapted performance on BioVid -- see Figure~\ref{fig:zero_kshot_by_setting}. In the binary task, the training configuration had virtually no effect on
either zero-shot accuracy (Friedman test, Holm-adjusted $p=1.000$, Kendall's $W=0.006$) or macro-F1 ($p=1.000$, $W=0.009$). In contrast, $k$-shot performance varied significantly across the training and
evaluation settings for both accuracy ($p<0.001$, $W=0.336$) and macro-F1 ($p<0.001$, $W=0.332$). The corresponding phase-by-setting interactions were also significant ($p<0.001$), indicating that episodic
task size primarily affected the benefit obtained from subject-specific support rather than the model's zero-shot capability.

A similar, but more nuanced, pattern was observed in the five-class task. Zero-shot accuracy remained stable across training configurations ($p=0.581$, $W=0.018$), whereas $k$-shot accuracy showed a large
setting effect ($p<0.001$, $W=0.609$). Macro-F1 varied across settings in both phases, although the effect was considerably larger for $k$-shot performance ($p<0.001$, $W=0.549$) than for zero-shot performance
($p<0.001$, $W=0.199$). Overall, these results suggest that the choice of episodic $k$ and $q$ mainly determines how effectively the model exploits target-subject support, while its unadapted accuracy remains largely insensitive to this choice, wehreas the F1-score is significantly impacted by this choice.

\subsubsection{Choice of Signal Channels}
\label{sec:signal_ablation}
Previous studies argued that the EDA channel in combination with the ECG channel carries the most information about the perceived pain of a participant. Table~\ref{tab:ablation_modality} shows the results of all possible combinations of two modalities in the binary classification tasks of the BioVid and the SenseEmotion datasets. The results confirm our assumption, as the combination of EDA and ECG yields the highest performance. However, the difference is marginal and not statistically significant.

\begin{table}[ht]
\small
    \centering
    \begin{tabular*}{\textwidth}{@{\extracolsep{\fill}}lcccc@{}}
        \toprule
        & \multicolumn{2}{c}{\textsc{BioVid}}
        & \multicolumn{2}{c}{\textsc{SenseEmotion}} \\
        \cmidrule(lr){2-3}
        \cmidrule(lr){4-5}
        \textbf{Modalities}
        & \textbf{Zero-Shot}
        & \textbf{$k$-Shot}
        & \textbf{Zero-Shot}
        & \textbf{$k$-Shot} \\
        \midrule

        EMG+RSP
        & N/A
        & N/A
        & 63.79\% $(\pm 11.6)$
        & 64.71\% $(\pm 11.7)$ \\

        ECG+RSP
        & N/A
        & N/A
        & 64.60\% $(\pm 11.3)$
        & 67.51\% $(\pm 12.2)$ \\

        EDA+RSP
        & N/A
        & N/A
        & 81.64\% $(\pm 10.4)$
        & 82.44\% $(\pm 10.7)$ \\
        
        EDA+EMG
        & 85.52\% $(\pm 9.31)$
        & 85.68\% $(\pm 9.31)$
        & 81.52\% $(\pm 9.31)$
        & 81.99\% $(\pm 10.7)$ \\

        ECG+EMG
        & 59.39\% $(\pm 15.2)$
        & 65.14\% $(\pm 16.0)$
        & 58.46\% $(\pm 10.1)$
        & 63.34\% $(\pm 11.8)$ \\

        EDA+ECG
        & \textbf{85.75\%} $(\pm 13.5)$
        & \textbf{86.25\%} $(\pm 14.2)$
        & \textbf{82.37\%} $(\pm 10.9)$
        & \textbf{83.43\%} $(\pm 11.2)$ \\

        \bottomrule
    \end{tabular*}

    \caption{Ablation study on the modalities used. In accordance with
    previous studies~\cite{Lu2023PainAttnNet,li2024automatic}, EDA+ECG achieves the highest classification accuracy in the
    binary ($T_0$ vs.\ $T_4$ and $T_0$ vs.\ $T_3$, respectively) LOSO setting.
    N/A indicates that the corresponding modality is not available in the dataset. We report accuracy and trained the models with $k,q=10$.}
    \label{tab:ablation_modality}
\end{table}

In summary, these ablation experiments suggest that the Few-Shot Subject-as-a-Task setup we devised in this work is
enabling the model to learn robust and transferable feature maps. To examine the effect of the support size \(k\), more experiments are needed where we keep \(q\) fixed and vary only \(k\).

\subsection{Experimental Setup}
\label{sec:expsetup}

We ran all our experiments on a single 16 GB consumer-grade GPU on Google Colab. We provide the hyper-parameters used
for training and evaluation in Table~\ref{tab:hyperparameters} and describe our architecture in Table~\ref{tab:can_architecture}.
The code for our implementation can be found under \url{https://github.com/hhihn/FewShotPainAdaptation}.

\begin{table}[ht]
  \centering
  \caption{Compact architecture of the proposed CAN model.}
  \label{tab:can_architecture}
  \setlength{\tabcolsep}{3pt}
  \renewcommand{\arraystretch}{1.15}
  \begin{tabular}{
    @{}
    p{0.19\columnwidth}
    p{0.48\columnwidth}
    p{0.23\columnwidth}
    @{}
  }
    \toprule
    \textbf{Stage} & \textbf{Core configuration} & \textbf{Output} \\
    \midrule

   Input
    & EDA and ECG windows with 1152 steps
    & $B\times1152\times2$ \\
\midrule
    Modality encoders
    & Independent temporal convolutional frontends with pooling
    & $2\times(B\times78\times16)$ \\
\midrule
    CrossMod fusion
    & Two 8-head self-attention layers followed by bidirectional
      cross-modal attention
    & $B\times78\times32$ \\
\midrule
    Prototype memory
    & \(M_{\mathrm{slot}}=2\) learned temporal prototype slots per class
    & $B_t\times|\mathcal{C}|M_{\mathrm{slot}}\times78\times32$ \\
\midrule
    Query encoding
    & The same modality encoders and CrossMod fusion are applied to each query
    & $B_t\times Q\times78\times32$ \\
\midrule
    CAN alignment
    & Pair-conditioned bidirectional temporal attention and attended cosine
      similarity, $\tau=0.025$
    & $B_t\times Q\times|\mathcal{C}|M_{\mathrm{slot}}$ \\
    \midrule
    Classification
    & Log-mean-exp over slots followed by non-negative logit scaling
    & $B_t\times Q\times|\mathcal{C}|$ \\
    \midrule
    Total Params. & Binary & 37.7K \\
     & Five Classes & 44.3K \\
    \bottomrule
  \end{tabular}
\end{table}

\begin{table}[ht]
  \centering
  \caption{Main hyperparameters used for the full LOSO experiment.}
  \label{tab:hyperparameters}
  \begin{tabular}{ll  }
    \toprule
    \textbf{Hyperparameter} & \textbf{Setting} \\
    \midrule
    Temporal frontend
    & Filters \(8/16\); kernels \(64/16\) \\

    Pooling / dropout / \(L_2\)
    & \(4,8\) / \(0.25\) / \(10^{-4}\) \\

    CrossMod configuration
    & 2 layers; 8 heads; hidden dim. 128 \\

    CAN configuration
    & \(\tau=0.025\); meta hidden dim. 32 \\

    CAN support source
    & Learned prototype memory \\

    Prototype memory
    & 2 slots/class; 256 init. samples/class \\

    Learning rate / schedule
    & \(6\times10^{-4}\) / cosine \\

    Training budget
    & 1 epoch; 20,000 tasks; batch size 16 \\

    Data augmentation
    & Gaussian noise \(0.01\); window shift \\

    Validation / held-out tasks
    & 50 / 500; checkpoint by F1 \\
    \bottomrule
  \end{tabular}
\end{table}

\section{Discussion}
\label{sec:discussion}
This study investigated whether Few-Shot learning can support personalised automated pain assessment.
With respect to the first research question, the Zero-Shot variant produced accuracies within the range of previously
reported LOSO results, while the support-conditioned $k$-Shot variant produced higher mean accuracy than the Zero-Shot
variant in every evaluated setting. The largest differences were observed in the multi-class tasks, for which the
$k$-Shot model reached 40.06\% accuracy on BioVid and 44.08\% on SenseEmotion. These values were numerically higher than
the best comparison values listed in Tables~\ref{tab:senseemotion_pain_recognition_performance},~\ref{tab:biovid_pain_recognition_performance}, and~\ref{tab:painmonit_pain_recognition_performance}, but they should not be interpreted as
evidence of superiority over those methods. In contrast to the comparison methods, $k$-Shot inference uses labelled
examples from the held-out subject, and the studies may also differ in preprocessing, model selection, and evaluation
details. The results instead show the performance attainable when a small, balanced set of subject-specific labels is
available. Additionally, any outperforming accuracy in the reported results in Tables~\ref{tab:senseemotion_pain_recognition_performance},~\ref{tab:biovid_pain_recognition_performance}, and~\ref{tab:painmonit_pain_recognition_performance} should be treated with care as no previous study reports statistical tests results comparing their achievement with previous study and all differences are marginal compared to the reported standard deviation of the LOSO setting. Thus, we focus more an how the $k$-Shot personalisation affects subject- and class-based results compared to the Zero-Shot case.

The subject-level analysis answers the second research question by showing that the association between personalisation
and performance is heterogeneous. Most held-out subjects had higher estimated macro-F1 in the $k$-Shot condition, and
the mean difference was positive, but a minority had no benefit or lower performance. The dependence on support-set size
is consistent with the hypothesis that a single atypical or noisy support window can yield an unrepresentative class
prototype, whereas several support samples provide a more stable estimate. This mechanism was not measured directly,
however, and the reported differences do not constitute fixed-query estimates of the causal effect of adding support:
Zero-Shot performance was computed on the complete held-out query set, whereas $k$-Shot performance was averaged over
episodes containing sampled support and query sets. A paired analysis using identical query samples before and after
adaptation is needed to isolate the effect of personalisation at subject level.

The class-wise findings address the third research question. Personalisation was associated with the largest F1 gains
for the intermediate BioVid pain levels, while class-0 F1 decreased and the maximum pain class showed a smaller gain.
Thus, the improvement in macro-F1 reflects a redistribution of errors rather than a uniform improvement across pain
levels. One possible explanation is that physiological responses to intermediate stimuli exhibit greater between-subject
variation or overlap more strongly than responses at the extremes, making subject-specific prototypes particularly
useful for these classes. The current experiments do not directly test this explanation. Relating class-wise gains to
within- and between-subject physiological variability, prototype dispersion, and changes in the confusion matrix would
help determine whether the observed pattern reflects improved subject calibration or a shift in the model's decision
boundaries that disadvantages the no-pain class.

These findings extend earlier work on personalised pain assessment and Few-Shot adaptation. Prior approaches have used
personalised physiological feature fusion and attention~\cite{jiang2024personalized}, source-free subject adaptation
from neutral facial data~\cite{sharafi2025disentangled}, participant selection~\cite{thiam2017hierarchical}, and
MAML-based adaptation of facial pain expressions~\cite{rathee2022pain}. The present results provide complementary
evidence for a prototype-based approach using ECG and EDA, with each held-out subject treated as a new task. In
particular, the prototype formulation permits adaptation without updating all model parameters. At the same time, the
need for labelled examples from every target class makes the $k$-Shot setting more demanding than approaches that adapt
from unlabelled data or from a neutral reference alone.

\subsection{Limitations}
Several limitations remain. First, the Zero-Shot
path uses transductive normalisation: its statistics are computed jointly from all query samples of the held-out
subject. This gives the model access to the distribution of the complete, although unlabelled, test batch and may not
represent deployment in which samples must be classified individually or arrive sequentially. Second, $k$-Shot
personalisation requires labelled target-subject examples for every class; its annotation cost therefore grows as
$N\times K$ and may be substantial in clinical settings, particularly when balanced examples of multiple pain levels
must be elicited. Third, the episodic estimates depend on randomly sampled support and query windows. Although results
are aggregated across 500 episodes, sampling variance and the use of different query sets across conditions limit
subject-level comparisons and prevent a fixed-query counterfactual interpretation. Fourth, training uses the tenfold
augmentation procedure of Thiam et al.~\cite{Thiam2021DeepPhys}, but we did not evaluate the method without augmentation or with
alternative augmentation schemes. The extent to which performance depends on this specific procedure is therefore
unknown, and augmented observations should not be interpreted as independent participants. Fifth, the reported results
do not quantify variation across independent training seeds, and the small numerical differences between methods in the
benchmark tables were not tested within a common experimental pipeline. Finally, BioVid, SenseEmotion, and PainMonit Experimental Dataset comprise
healthy participants exposed to controlled heat stimuli, and our experiments use only ECG and EDA. The labels therefore
represent experimentally administered stimulus levels rather than clinical pain assessments or self-reported pain.
Validation in clinical populations and under less controlled acquisition conditions is required before the findings can
be generalised to other pain aetiologies, sensor configurations, or deployment environments.

\subsection{Future Work}
Future work should first evaluate personalisation with a paired, fixed-query protocol and repeated training seeds. The
methodological contribution can then be isolated through comparisons of multiple prototype slots with one prototype per
class, the learned prototype bank with a population-mean prototype, and support-based normalisation with fixed
training-population normalisation. Prototype selection and weighting could also be learned through hierarchical
information fusion~\cite{hihn2020hierarchical}, while continual-learning mechanisms could update subject prototypes as
new labelled observations become available~\cite{hihn2021mixture,hihn2023hierarchically,hihn2024online}. For practical
deployment, future studies should reduce the requirement for balanced labels from every pain level, support sequential
rather than transductive inference, and evaluate calibration, macro-F1, and clinically relevant error costs in addition
to accuracy.

\subsection{Conclusion}
In summary, we introduced and evaluated subject-as-a-task Few-Shot learning for personalised classification of
experimentally induced heat-pain levels. Under the present episodic protocol, subject-specific support was associated
with higher mean performance, particularly in the multi-class tasks and for intermediate pain levels, but the benefit was
not uniform across subjects or classes. The Zero-Shot variant requires no labelled target-subject samples, although its
current transductive normalisation assumes access to an unlabelled batch from that subject. These findings support further
investigation of prototype-based personalisation, but fixed-query comparisons and validation in clinical settings are
needed before conclusions can be drawn about individualised pain monitoring in practice.

\section*{Acknowledgements}
This work was financially supported by funds under the IU Incubator programme \textit{TAPAS: Tailored Automated Pain Assessment Systems}.

\section*{Author Contributions}
\textbf{Conceptualization:} Heinke Hihn \newline
\textbf{Formal Analysis:} Heinke Hihn \newline
\textbf{Investigation:} Heinke Hihn \newline
\textbf{Resources:} Heinke Hihn \newline
\textbf{Software:} Heinke Hihn, Ibrahim Eisawy \newline
\textbf{Supervision:} Friedhelm Schwenker, Hans A. Kestler \newline
\textbf{Data curation}: Heinke Hihn, Patrick Thiam \newline
\textbf{Visualization:} Heinke Hihn \newline
\textbf{Funding acquisition:} Heinke Hihn \newline
\textbf{Writing - original draft:} Heinke Hihn \newline
\textbf{Writing - review and editing:} Heinke Hihn, Ibrahim Eisawy, Patrick Thiam, Friedhelm Schwenker, Hans A. Kestler

\bibliography{main}

@inproceedings{velana2016senseemotion,
  title={The senseemotion database: A multimodal database for the development and systematic validation of an automatic pain-and emotion-recognition system},
  author={Velana, Maria and Gruss, Sascha and Layher, Georg and Thiam, Patrick and Zhang, Yan and Schork, Daniel and Kessler, Viktor and Meudt, Sascha and Neumann, Heiko and Kim, Jonghwa and others},
  booktitle={IAPR Workshop on Multimodal Pattern Recognition of Social Signals in Human-Computer Interaction},
  pages={127--139},
  year={2016},
  organization={Springer}
}

@article{anter2023rlwoa,
  title={RLWOA-SOFL: A new learning model-based reinforcement swarm intelligence and self-organizing deep fuzzy rules for fMRI pain decoding},
  author={Anter, Ahmed M and Zhang, Zhiguo},
  journal={IEEE Transactions on Affective Computing},
  volume={15},
  number={2},
  pages={644--656},
  year={2023},
  publisher={IEEE}
}

@article{khan2026chronic,
  title={Chronic Pain as a Continuum: Unsupervised Learning for Identification and Quantification of Coexisting Chronic Pain Mechanisms},
  author={Khan, Md Asif and Koh, Ryan GL and Kumbhare, Dinesh and Doyle, Thomas E},
  journal={IEEE Access},
  year={2026},
  publisher={IEEE}
}

@inproceedings{Walter2013BioVid,
  author = {Steffen Walter and Sascha Gruss and Hagen Ehleiter and Junwen Tan and Harald C. Traue and Stephen Crawcour and Philipp Werner and Ayoub Al-Hamadi and Adriano O. Andrade and Gustavo Moreira da Silva},
  title = {The biovid heat pain database: Data for the advancement and systematic validation of an automated pain recognition system},
  booktitle = {2013 IEEE International Conference on Cybernetics (CYBCO)},
  pages = {128-131},
  year = {2013},
  doi = {10.1109/CYBConf.2013.6617456}
}

@article{CrossModTransformer2025,
  author = {Jaleh Farmani and Ghazal Bargshady and Stefanos Gkikas and Manolis Tsiknakis and Raul Fernandez Rojas},
  title = {CrossMod-Transformer: multi-modal pain detection through EDA and ECG fusion},
  journal = {Scientific Reports},
  volume = {15},
  number = {1},
  pages = {29467},
  year = {2025},
  doi = {10.1038/s41598-025-14238-y}
}

@article{Lu2023PainAttnNet,
  author = {Zhenyuan Lu and Burcu Ozek and Sagar Kamarthi},
  title = {Transformer encoder with multiscale deep learning for pain classification using physiological signals},
  journal = {Frontiers in Physiology},
  volume = {14},
  pages = {1294577},
  year = {2023},
  doi = {10.3389/fphys.2023.1294577}
}

@Article{ricken2023pain,
AUTHOR = {Ricken, Tobias Benjamin and Bellmann, Peter and Walter, Steffen and Schwenker, Friedhelm},
TITLE = {Pain Detection in Biophysiological Signals: Knowledge Transfer from Short-Term to Long-Term Stimuli Based on Distance-Specific Segment Selection},
JOURNAL = {Computers},
VOLUME = {12},
YEAR = {2023},
NUMBER = {4},
ARTICLE-NUMBER = {71},
URL = {https://www.mdpi.com/2073-431X/12/4/71},
ISSN = {2073-431X},
DOI = {10.3390/computers12040071}
}

@inproceedings{li2024automatic,
  title={Automatic pain assessment based on physiological signals: Application of multi-scale networks and cross-attention cross-attention},
  author={Li, JiaHao and Luo, JinCheng and Wang, YanSheng and Jiang, YunXiang and Chen, Xu and Quan, YuJuan},
  booktitle={Proceedings of the 2024 13th International Conference on Bioinformatics and Biomedical Science},
  pages={113--122},
  year={2024}
}

@article{thiam2019exploring,
  title={Exploring deep physiological models for nociceptive pain recognition},
  author={Thiam, Patrick and Bellmann, Peter and Kestler, Hans A and Schwenker, Friedhelm},
  journal={Sensors},
  volume={19},
  number={20},
  pages={4503},
  year={2019},
  publisher={MDPI}
}

@article{Thiam2021DeepPhys,
  author = {Patrick Thiam and Heinke Hihn and Daniel A. Braun and Hans A. Kestler and Friedhelm Schwenker},
  title = {Multi-modal pain intensity assessment based on physiological signals: a deep learning perspective},
  journal = {Frontiers in Physiology},
  volume = {12},
  pages = {720464},
  year = {2021},
  doi = {10.3389/fphys.2021.720464} 
}

@inproceedings{fang2022atlas,
  title={Atlas: An adaptive transfer learning based pain assessment system: A real life unsupervised pain assessment solution},
  author={Fang, Ruijie and Zhang, Ruoyu and Hosseini, Elahe and Orooji, Mahdi and Homayoun, Houman and Hosseini, Sayed Mohammad and Faghih, Mahya and Rafatirad, Soheil and Rafatirad, Setareh},
  booktitle={2022 44th Annual International Conference of the IEEE Engineering in Medicine \& Biology Society (EMBC)},
  pages={1331--1337},
  year={2022},
  organization={IEEE}
}

@inproceedings{sharafi2025disentangled,
  title={Disentangled source-free personalization for facial expression recognition with neutral target data},
  author={Sharafi, Masoumeh and Ollivier, Emma and Zeeshan, Muhammad Osama and Belharbi, Soufiane and Koerich, Alessandro Lameiras and Pedersoli, Marco and Bacon, Simon and Granger, Eric},
  booktitle={2025 IEEE 19th International Conference on Automatic Face and Gesture Recognition (FG)},
  pages={1--10},
  year={2025},
  organization={IEEE}
}

@article{Aziz2024MDNet,
  author = {Sumair Aziz and Muttaba Ur Khan and Raul Fernandez Rojas},
  title = {MDNet: A Lightweight Multidomain 1-D CNN for Embedded Pain Assessment Using EDA Signals},
  journal = {IEEE Sensors Journal},
  volume = {24},
  number = {11},
  pages = {17637-17646},
  year = {2024},
  doi = {10.1109/JSEN.2024.3385802}
}

@article{Pouromran2021Exploration,
  author = {Fatemeh Pouromran and Srinivasan Radhakrishnan and Sagar Kamarthi},
  title = {Exploration of physiological sensors, features, and machine learning models for pain intensity estimation},
  journal = {PLOS ONE},
  volume = {16},
  number = {7},
  pages = {e0254108},
  year = {2021},
  doi = {10.1371/journal.pone.0254108}
}

@article{Steur2025Supervised,
  author = {Steur, N.A.K. and Schwenker, F.},
  title = {Multimodal Pain Recognition Based on Contrastive Adversarial Autoencoder Pretraining},
  journal = {Machine Learning and Knowledge Extraction},
  volume = {7},
  pages = {165},
  year = {2025},
  doi = {10.3390/make7040165}
}

@article{hou2019cross,
  title={Cross attention network for few-shot classification},
  author={Hou, Ruibing and Chang, Hong and Ma, Bingpeng and Shan, Shiguang and Chen, Xilin},
  journal={Advances in neural information processing systems},
  volume={32},
  year={2019}
}

@article{gouverneur2021comparison,
  title={Comparison of feature extraction methods for physiological signals for heat-based pain recognition},
  author={Gouverneur, Philip and Li, Fr{\'e}d{\'e}ric and Adamczyk, Wac{\l}aw M and Szikszay, Tibor M and Luedtke, Kerstin and Grzegorzek, Marcin},
  journal={Sensors},
  volume={21},
  number={14},
  pages={4838},
  year={2021},
  publisher={MDPI}
}

@article{gouverneur2023explainable,
  title={Explainable artificial intelligence (XAI) in pain research: understanding the role of electrodermal activity for automated pain recognition},
  author={Gouverneur, Philip and Li, Fr{\'e}d{\'e}ric and Shirahama, Kimiaki and Luebke, Luisa and Adamczyk, Wac{\l}aw M and Szikszay, Tibor M and Luedtke, Kerstin and Grzegorzek, Marcin},
  journal={Sensors},
  volume={23},
  number={4},
  pages={1959},
  year={2023},
  publisher={MDPI}
}

@inproceedings{gkikas2025efficient,
  title={Efficient pain recognition via respiration signals: A single cross-attention transformer multi-window fusion pipeline},
  author={Gkikas, Stefanos and Kyprakis, Ioannis and Tsiknakis, Manolis},
  booktitle={Companion Proceedings of the 27th International Conference on Multimodal Interaction},
  pages={70--79},
  year={2025}
}

@article{gouverneur2024experimental,
  title={An experimental and clinical physiological signal dataset for automated pain recognition},
  author={Gouverneur, Philip and Badura, Aleksandra and Li, Fr{\'e}d{\'e}ric and Bie{\'n}kowska, Maria and Luebke, Luisa and Adamczyk, Wac{\l}aw M and Szikszay, Tibor M and My{\'s}liwiec, Andrzej and Luedtke, Kerstin and Grzegorzek, Marcin and others},
  journal={Scientific Data},
  volume={11},
  number={1},
  pages={1051},
  year={2024},
  publisher={Nature Publishing Group UK London}
}

@article{luebke2023objective,
  title={Objective measurement of subjective pain perception with autonomic body reactions in healthy subjects and chronic back pain patients: an experimental heat pain study},
  author={Luebke, Luisa and Gouverneur, Philip and Szikszay, Tibor M and Adamczyk, Wac{\l}aw M and Luedtke, Kerstin and Grzegorzek, Marcin},
  journal={Sensors},
  volume={23},
  number={19},
  pages={8231},
  year={2023},
  publisher={MDPI}
}

@inproceedings{shi2022tree,
  title={Tree-based models for pain detection from biomedical signals},
  author={Shi, Heng and Chikhaoui, Belkacem and Wang, Shengrui},
  booktitle={International conference on smart homes and health telematics},
  pages={183--195},
  year={2022},
  organization={Springer}
}

@article{werner2019automatic,
  title={Automatic recognition methods supporting pain assessment: A survey},
  author={Werner, Philipp and Lopez-Martinez, Daniel and Walter, Steffen and Al-Hamadi, Ayoub and Gruss, Sascha and Picard, Rosalind W},
  journal={IEEE Transactions on Affective Computing},
  volume={13},
  number={1},
  pages={530--552},
  year={2019},
  publisher={IEEE}
}

@inproceedings{akiba2019optuna,
  title={{O}ptuna: A Next-Generation Hyperparameter Optimization Framework},
  author={Akiba, Takuya and Sano, Shotaro and Yanase, Toshihiko and Ohta, Takeru and Koyama, Masanori},
  booktitle={The 25th ACM SIGKDD International Conference on Knowledge Discovery \& Data Mining},
  pages={2623--2631},
  year={2019}
}

@article{wang2019survey,
  title={A survey of zero-shot learning: Settings, methods, and applications},
  author={Wang, Wei and Zheng, Vincent W and Yu, Han and Miao, Chunyan},
  journal={ACM Transactions on Intelligent Systems and Technology (TIST)},
  volume={10},
  number={2},
  pages={1--37},
  year={2019},
  publisher={ACM New York, NY, USA}
}

@article{kachele2016methods,
  title={Methods for person-centered continuous pain intensity assessment from bio-physiological channels},
  author={K{\"a}chele, Markus and Thiam, Patrick and Amirian, Mohammadreza and Schwenker, Friedhelm and Palm, G{\"u}nther},
  journal={IEEE Journal of Selected Topics in Signal Processing},
  volume={10},
  number={5},
  pages={854--864},
  year={2016},
  publisher={IEEE}
}

@article{jiang2024personalized,
  title={Personalized and adaptive neural networks for pain detection from multi-modal physiological features},
  author={Jiang, Mingzhe and Rosio, Riitta and Salanter{\"a}, Sanna and Rahmani, Amir M and Liljeberg, Pasi and da Silva, Daniel S and de Albuquerque, Victor Hugo C and Wu, Wanqing},
  journal={Expert Systems with Applications},
  volume={235},
  pages={121082},
  year={2024},
  publisher={Elsevier}
}

@inproceedings{aslam2025learning,
  title={Learning from stochastic teacher representations using student-guided knowledge distillation},
  author={Aslam, Muhammad Haseeb and Martinez, Clara and Pedersoli, Marco and Koerich, Alessandro Lameiras and Etemad, Ali and Granger, Eric},
  booktitle={Joint European Conference on Machine Learning and Knowledge Discovery in Databases},
  pages={235--253},
  year={2025},
  organization={Springer}
}

@article{hawker2011measures,
  title={Measures of adult pain: Visual analog scale for pain (vas pain), numeric rating scale for pain (nrs pain), mcgill pain questionnaire (mpq), short-form mcgill pain questionnaire (sf-mpq), chronic pain grade scale (cpgs), short form-36 bodily pain scale (sf-36 bps), and measure of intermittent and constant osteoarthritis pain (icoap)},
  author={Hawker, Gillian A and Mian, Samra and Kendzerska, Tetyana and French, Melissa},
  journal={Arthritis care \& research},
  volume={63},
  number={S11},
  pages={S240--S252},
  year={2011},
  publisher={John Wiley \& Sons, Inc. Hoboken, USA}
}

@article{belal2025fsid,
  title={FSID: a novel approach to human activity recognition using few-shot weight imprinting},
  author={Belal, Mohammad and Hassan, Taimur and Hassan, Abdelfatah and Velayudhan, Divya and Elhendawi, Noureldin and Aljarah, Ahmad and Hussain, Irfan},
  journal={Scientific Reports},
  volume={15},
  number={1},
  pages={20865},
  year={2025},
  publisher={Nature Publishing Group UK London}
}

@article{an2023dual,
  title={Dual attention relation network with fine-tuning for few-shot EEG motor imagery classification},
  author={An, Sion and Kim, Soopil and Chikontwe, Philip and Park, Sang Hyun},
  journal={IEEE transactions on neural networks and learning systems},
  volume={35},
  number={11},
  pages={15479--15493},
  year={2023},
  publisher={IEEE}
}

@article{wang2026msfsnet,
  title={MSFSNet: Multi-Source Few-Shot Adaptation Network for Cross-Subject Depression Recognition from EEG Signals},
  author={Wang, Kang and Zhang, Yanan and Zhang, Yingwei and Zhang, Fa and Shen, Jian and Hu, Bin},
  journal={IEEE Journal of Biomedical and Health Informatics},
  year={2026},
  publisher={IEEE}
}

@article{rathee2022pain,
  title={Pain detection from facial expressions using domain adaptation technique},
  author={Rathee, Neeru and Pahal, Sudesh and Sheoran, Poonam},
  journal={Pattern Analysis and Applications},
  volume={25},
  number={3},
  pages={567--574},
  year={2022},
  publisher={Springer}
}

@article{buzuti2024generative,
  title={Generative AI for neonatal pain assessment: a sound approach to improve data-driven automatic recognition in intensive care unit},
  author={Buzuti, Lucas and Giraldi, Gilson and Heiderich, Tatiany and Barros, Marina and Guinsburg, Ruth and Thomaz, Carlos},
  journal={Available at SSRN 4994076},
  year={2024}
}

@inproceedings{mccofie2025few,
  title={Few-Shot Prompting with Vision Language Model for Pain Classification in Infant Cry Sounds},
  author={McCofie, Anthony and Kandiyana, Abhiram and Mouton, Peter R and Sun, Yu and Goldgof, Dmitry},
  booktitle={2025 IEEE 38th International Symposium on Computer-Based Medical Systems (CBMS)},
  pages={857--862},
  year={2025},
  organization={IEEE}
}

@article{el2026pandia,
  title={PANDIA: Personalized neuro-symbolic multimodal fusion for interpretable neonatal pain assessment},
  author={El Othmani, Oussama and Naouali, Sami},
  journal={PLOS Digital Health},
  volume={5},
  number={5},
  pages={e0001442},
  year={2026},
  publisher={Public Library of Science San Francisco, CA USA}
}

@article{khan2025systematic,
  title={A Systematic Review of Multimodal Signal Fusion for Acute Pain Assessment Systems},
  author={Khan, Muhammad Umar and Chetty, Girija and Goecke, Roland and Fernandez-Rojas, Raul},
  journal={ACM Computing Surveys},
  volume={58},
  number={2},
  pages={1--35},
  year={2025},
  publisher={ACM New York, NY}
}

@inproceedings{ji2024unraveling,
  title={Unraveling pain levels: A data-uncertainty guided approach for effective pain assessment},
  author={Ji, Xinwei and Chang, Xiaomin and Li, Wei and Zomaya, Albert Y},
  booktitle={Proceedings of the AAAI Conference on Artificial Intelligence},
  volume={38},
  number={20},
  pages={22167--22175},
  year={2024}
}

@article{werner2016automatic,
  title={Automatic pain assessment with facial activity descriptors},
  author={Werner, Philipp and Al-Hamadi, Ayoub and Limbrecht-Ecklundt, Kerstin and Walter, Steffen and Gruss, Sascha and Traue, Harald C},
  journal={IEEE Transactions on Affective Computing},
  volume={8},
  number={3},
  pages={286--299},
  year={2016},
  publisher={IEEE}
}

@inproceedings{lopez2018continuous,
  title={Continuous pain intensity estimation from autonomic signals with recurrent neural networks},
  author={Lopez-Martinez, Daniel and Picard, Rosalind},
  booktitle={2018 40th Annual International Conference of the IEEE Engineering in Medicine and Biology Society (EMBC)},
  pages={5624--5627},
  year={2018},
  organization={IEEE}
}

@article{hospedales2021meta,
  title={Meta-learning in neural networks: A survey},
  author={Hospedales, Timothy and Antoniou, Antreas and Micaelli, Paul and Storkey, Amos},
  journal={IEEE transactions on pattern analysis and machine intelligence},
  volume={44},
  number={9},
  pages={5149--5169},
  year={2021},
  publisher={IEEE}
}

@article{gui2024survey,
  title={A survey on self-supervised learning: Algorithms, applications, and future trends},
  author={Gui, Jie and Chen, Tuo and Zhang, Jing and Cao, Qiong and Sun, Zhenan and Luo, Hao and Tao, Dacheng},
  journal={IEEE Transactions on Pattern Analysis and Machine Intelligence},
  volume={46},
  number={12},
  pages={9052--9071},
  year={2024},
  publisher={IEEE}
}

@inproceedings{gkikas2021multi,
  title={Multi-task neural networks for pain intensity estimation using electrocardiogram and demographic factors},
  author={Gkikas, Stefanos and Chatzaki, Chariklia and Tsiknakis, Manolis},
  booktitle={International Conference on Information and Communication Technologies for Ageing Well and e-Health},
  pages={324--337},
  year={2021},
  organization={Springer}
}

@inproceedings{werner2014automatic,
  title={Automatic pain recognition from video and biomedical signals},
  author={Werner, Philipp and Al-Hamadi, Ayoub and Niese, Robert and Walter, Steffen and Gruss, Sascha and Traue, Harald C},
  booktitle={2014 22nd international conference on pattern recognition},
  pages={4582--4587},
  year={2014},
  organization={IEEE}
}

@inproceedings{lopez2017multi,
  title={Multi-task neural networks for personalized pain recognition from physiological signals},
  author={Lopez-Martinez, Daniel and Picard, Rosalind},
  booktitle={2017 Seventh International Conference on Affective Computing and Intelligent Interaction Workshops and Demos (ACIIW)},
  pages={181--184},
  year={2017},
  organization={IEEE}
}

@article{wang2020generalizing,
  title={Generalizing from a few examples: A survey on few-shot learning},
  author={Wang, Yaqing and Yao, Quanming and Kwok, James T and Ni, Lionel M},
  journal={ACM computing surveys (csur)},
  volume={53},
  number={3},
  pages={1--34},
  year={2020},
  publisher={ACM New York, NY, USA}
}

@article{gozzi2024unraveling,
  title={Unraveling the physiological and psychosocial signatures of pain by machine learning},
  author={Gozzi, Noemi and Preatoni, Greta and Ciotti, Federico and Hubli, Mich{\`e}le and Schweinhardt, Petra and Curt, Armin and Raspopovic, Stanisa},
  journal={Med},
  volume={5},
  number={12},
  pages={1495--1509},
  year={2024},
  publisher={Elsevier}
}

@inproceedings{bourou2018video,
  title={Video-based pain level assessment: feature selection and inter-subject variability modeling},
  author={Bourou, Dimitra and Pampouchidou, Anastasia and Tsiknakis, Manolis and Marias, Kostas and Simos, Panagiotis},
  booktitle={2018 41st International Conference on Telecommunications and Signal Processing (TSP)},
  pages={1--6},
  year={2018},
  organization={IEEE}
}

@article{thiam2025dealing,
  title={Dealing with Class Overlap Through Cluster-Based Sample Weighting},
  author={Thiam, Patrick and Schwenker, Friedhelm and Kestler, Hans Armin},
  journal={Computers},
  volume={14},
  number={11},
  pages={457},
  year={2025},
  publisher={MDPI}
}

@article{hihn2020specialization,
  title={Specialization in hierarchical learning systems: A unified information-theoretic approach for supervised, unsupervised and reinforcement learning},
  author={Hihn, Heinke and Braun, Daniel A},
  journal={Neural Processing Letters},
  volume={52},
  number={3},
  pages={2319--2352},
  year={2020},
  publisher={Springer}
}

@article{hihn2026efficient,
      title={Efficient Architecture Search under Leave-One-Subject-Out Evaluation}, 
      author={Hihn, Heinke and Schwenker, Friedhelm},
      year={2026},
      eprint={2609.21457},
      archivePrefix={arXiv},
      primaryClass={cs.LG},
      journal={arXiv},
      url={https://arxiv.org/abs/2609.21457}, 
}

@article{li2021diversity,
  title={On the diversity of multi-head attention},
  author={Li, Jian and Wang, Xing and Tu, Zhaopeng and Lyu, Michael R},
  journal={Neurocomputing},
  volume={454},
  pages={14--24},
  year={2021},
  publisher={Elsevier}
}

@article{thuseethan2021deep,
  title={Deep continual learning for emerging emotion recognition},
  author={Thuseethan, Selvarajah and Rajasegarar, Sutharshan and Yearwood, John},
  journal={IEEE Transactions on Multimedia},
  volume={24},
  pages={4367--4380},
  year={2021},
  publisher={IEEE}
}

@article{wang2024comprehensive,
  title={A comprehensive survey of continual learning: Theory, method and application},
  author={Wang, Liyuan and Zhang, Xingxing and Su, Hang and Zhu, Jun},
  journal={IEEE transactions on pattern analysis and machine intelligence},
  volume={46},
  number={8},
  pages={5362--5383},
  year={2024},
  publisher={IEEE}
}

@article{del2023applications,
  title={Applications of self-supervised learning to biomedical signals: A survey},
  author={Del Pup, Federico and Atzori, Manfredo},
  journal={IEEE Access},
  volume={11},
  pages={144180--144203},
  year={2023},
  publisher={IEEE}
}

@article{hihn2024online,
  title={Online continual learning through unsupervised mutual information maximization},
  author={Hihn, Heinke and Braun, Daniel A},
  journal={Neurocomputing},
  volume={578},
  pages={127422},
  year={2024},
  publisher={Elsevier}
}

@inproceedings{hihn2021mixture,
  title={Mixture-of-Variational-Experts for Continual Learning},
  author={Hihn, Heinke and Braun, Daniel Alexander},
  year={2021},
  booktitle={ICLR Workshop on Agent Learning in Open-Endedness}
}

@article{hihn2023hierarchically,
  title={Hierarchically structured task-agnostic continual learning},
  author={Hihn, Heinke and Braun, Daniel A},
  journal={Machine Learning},
  volume={112},
  number={2},
  pages={655--686},
  year={2023},
  publisher={Springer}
}

@inproceedings{thiam2017hierarchical,
  title={Hierarchical Combination of Video Features for Personalised Pain Level Recognition.},
  author={Thiam, Patrick and Kessler, Viktor and Schwenker, Friedhelm},
  booktitle={ESANN},
  pages={465--470},
  year={2017}
}

@inproceedings{hihn2020hierarchical,
  title={Hierarchical Expert Networks for Meta-Learning},
  author={Hihn, Heinke and Braun, Daniel A},
  booktitle={4th Lifelong Machine Learning Workshop at ICML 2020},
  year={2020}
}

@article{thiam2019multi,
  title={Multi-modal pain intensity recognition based on the senseemotion database},
  author={Thiam, Patrick and Kessler, Viktor and Amirian, Mohammadreza and Bellmann, Peter and Layher, Georg and Zhang, Yan and Velana, Maria and Gruss, Sascha and Walter, Steffen and Traue, Harald C and others},
  journal={IEEE Transactions on Affective Computing},
  volume={12},
  number={3},
  pages={743--760},
  year={2019},
  publisher={IEEE}
}

@article{thiam2021multi,
  title={Multi-modal pain intensity assessment based on physiological signals: A deep learning perspective},
  author={Thiam, Patrick and Hihn, Heinke and Braun, Daniel A and Kestler, Hans A and Schwenker, Friedhelm},
  journal={Frontiers in Physiology},
  volume={12},
  pages={720464},
  year={2021},
  publisher={Frontiers Media SA}
}

@article{thiam2020two,
  title={Two-stream attention network for pain recognition from video sequences},
  author={Thiam, Patrick and Kestler, Hans A and Schwenker, Friedhelm},
  journal={Sensors},
  volume={20},
  number={3},
  pages={839},
  year={2020},
  publisher={MDPI}
}

@inproceedings{kessler2017multimodal,
  title={Multimodal fusion including camera photoplethysmography for pain recognition},
  author={Kessler, Viktor and Thiam, Patrick and Amirian, Mohammadreza and Schwenker, Friedhelm},
  booktitle={2017 International Conference on Companion Technology (ICCT)},
  pages={1--4},
  year={2017},
  organization={IEEE}
}

\end{document}